\documentclass{article}

\usepackage{microtype}
\usepackage{graphicx}
\usepackage{subcaption}
\usepackage{booktabs} 

\usepackage{hyperref}
\usepackage{bookmark}

\usepackage[preprint]{icml2026}
\makeatletter
\renewcommand{\Notice@String}{}
\makeatother

\usepackage{lmodern}

\usepackage{amsmath}
\usepackage{amssymb}
\usepackage{mathtools}
\usepackage{amsthm}
\usepackage{wrapfig}

\usepackage{url}

\usepackage{enumitem}
\usepackage{float}

\usepackage{algorithm}
\renewcommand{\algorithmicrequire}{\textbf{Input:}}
\renewcommand{\algorithmicensure}{\textbf{Output:}}
\usepackage{multirow}
\usepackage{makecell}
\usepackage[capitalize,noabbrev]{cleveref}

\theoremstyle{plain}

\theoremstyle{definition}

\theoremstyle{remark}

\usepackage[disable,textsize=tiny]{todonotes}

\usepackage{xcolor}
\newcommand{\midprice}[1][t]{p^{\text{mid}}_{#1}}
\newcommand{\midpriceest}[1][t]{\hat{p}^{\text{mid}}_{#1}}
\newcommand{\execprice}[1][t]{p^{\text{exec}}_{#1}}
\newcommand{\orderprice}[1][t]{p^{\text{order}}_{#1}}

\icmltitlerunning{TradeFM}

\begin{document}
\raggedbottom

\onecolumn

\icmltitle{TradeFM: A Generative Foundation Model for \\ Trade-flow and Market Microstructure}

\icmlsetsymbol{equal}{*}

\begin{icmlauthorlist}
  \icmlauthor{Maxime Kawawa-Beaudan}{equal}
  \icmlauthor{Srijan Sood}{equal} \\
  \icmlauthor{Kassiani Papasotiriou}{}
  \icmlauthor{Daniel Borrajo}{}
  \icmlauthor{Manuela Veloso}{}
\end{icmlauthorlist}

\vskip 0.1in
\begin{center}
  J.P. Morgan AI Research, New York, NY, USA
\end{center}

\icmlcorrespondingauthor{Srijan Sood,}{\texttt{\{first\_name\}.\{last\_name\}@jpmorgan.com}}

\icmlkeywords{Foundation Model,Generative Model,Market Microstructure,Algorithmic Trading,Trade-flow,Financial Time Series,Reinforcement Learning,Multi-Agent Simulation,Transformer,TradeFM}

\vskip 0.2in

\makeatletter
\gdef\@icmltitlerunning{TradeFM: A Generative Foundation Model for Trade-flow and Market Microstructure}
\renewcommand{\printAffiliationsAndNotice}[1]{\global\icml@noticeprintedtrue%
  \stepcounter{@affiliationcounter}%
  {\let\thefootnote\relax\footnotetext{\raggedright\hspace*{-\footnotesep}\ificmlshowauthors #1\fi%
      \forloop{@affilnum}{1}{\value{@affilnum} < \value{@affiliationcounter}}{%
        \textsuperscript{\arabic{@affilnum}}\ifcsname @affilname\the@affilnum\endcsname%
          \csname @affilname\the@affilnum\endcsname%
        \else
          {\bf AUTHORERR: Missing \textbackslash{}icmlaffiliation.}%
        \fi
      }%
      \ifdefined\icmlcorrespondingauthor@text
         \unskip.\ Correspondence to: \icmlcorrespondingauthor@text.%
      \else
        {\bf AUTHORERR: Missing \textbackslash{}icmlcorrespondingauthor.}%
      \fi
      \ \\
      \Notice@String
    }
  }
}
\makeatother
\printAffiliationsAndNotice{\icmlEqualContribution}

\begin{abstract}
  Foundation models have transformed domains from language to genomics by learning general-purpose representations from large-scale, heterogeneous data.
We introduce TradeFM, a 524M-parameter generative Transformer that brings this paradigm to market microstructure, learning directly from billions of trade events across $>$9K equities.
To enable cross-asset generalization, we develop scale-invariant features and a universal tokenization scheme that map the heterogeneous, multi-modal event stream of order flow into a unified discrete sequence -- eliminating asset-specific calibration.
Integrated with a deterministic market simulator, TradeFM-generated rollouts reproduce key stylized facts of financial returns, including heavy tails, volatility clustering, and absence of return autocorrelation.
Quantitatively, TradeFM achieves 2--3$\times$ lower distributional error than Compound Hawkes baselines and generalizes zero-shot to geographically out-of-distribution APAC markets with moderate perplexity degradation.
Together, these results suggest that scale-invariant trade representations capture transferable structure in market microstructure, opening a path toward synthetic data generation, stress testing, and learning-based trading agents.

\end{abstract}

\section{Introduction} \label{sec:introduction}
Financial markets are complex adaptive systems where the strategic interactions of heterogeneous participants give rise to high-frequency, non-stationary dynamics~\citep{bouchaud2010endogenous}.
The atomic record of these interactions is \textbf{order flow}: the stream of buy and sell orders submitted to the market~\citep{sirignano2021universal}.
Modeling order flow is a formidable challenge -- the statistical properties of trade data vary dramatically across assets, liquidity regimes, and time periods~\citep{Pasca2015151167}, and any single participant observes only a partial view of the true market state.

Despite this complexity, there is strong evidence that price formation follows \textit{universal} principles across markets.
\citet{sirignano2021universal} showed that a single deep learning model trained on pooled multi-stock data outperforms asset-specific models, even for held-out stocks.
This finding motivates a natural question: \textit{can a single foundation model learn a general-purpose representation of market mechanics from raw, multi-asset order flow?}

We answer in the affirmative with \textbf{TradeFM}, a 524M-parameter decoder-only Transformer pre-trained on over 10 billion tokens from $>$9K US equities.
Our contributions are fourfold:
\begin{enumerate}
    \item \textbf{TradeFM}: A large-scale generative foundation model for market microstructure that learns unified trade-flow dynamics from billions of transactions across the breadth of the US equity market.
    
    \item \textbf{Learning from Partial Observations}: Unlike prior deep learning approaches that require the full limit order book as input (Table~\ref{tab:related_work_comparison}), TradeFM learns from a partially observed market state – the event stream available to any single participant – demonstrating that a foundation model for microstructure does not require privileged access to the full order book.
    
    \item \textbf{Scale-Invariant Representation and Tokenization}: An end-to-end methodology that transforms raw, heterogeneous trade data into a unified discrete sequence via scale-invariant features and a universal tokenization scheme, enabling a single model to generalize across diverse assets and liquidity regimes without asset-specific calibration.
    
    \item \textbf{Closed-Loop Market Simulation}: Integration of the pre-trained model with a deterministic market simulator, creating a high-fidelity environment for evaluating realism via stylized fact reproduction, studying market impact, and training learning-based agents.
\end{enumerate}

In closed-loop evaluation, TradeFM-generated rollouts reproduce canonical stylized facts of financial returns.
Quantitatively, it achieves 2--3$\times$ lower distributional error than Compound Hawkes baselines and generalizes zero-shot to out-of-distribution APAC markets with moderate perplexity degradation.

\section{Background}\label{sec:background}

\subsection{The Mechanics of Modern Electronic Markets}\label{sec:background/definitions}
To provide context for a general AI/ML audience, we briefly introduce the core concepts of market microstructure fundamental to this work, which are standard features of modern electronic markets~\citep{hasbrouck2007empirical}.

Financial markets are predominantly organized around a \textbf{Limit Order Book (LOB)}, a real-time record of all outstanding orders for a security that functions as a continuous, double-sided auction. It consists of a \textbf{bid} (buy) side and an \textbf{ask} (sell) side; the midpoint between the highest bid and the lowest ask is an asset's \textbf{mid-price}. The ease with which an asset can be bought or sold quickly at a stable price is the asset's \textbf{liquidity}.

Market participants interact with the LOB through a sequence of actions, collectively known as \textbf{order flow}. Participants may submit \textbf{limit orders} with a specific price limit, which sit on the book waiting to be matched. The distance between the \textbf{order price} and the midprice is the \textbf{price depth}, quoted in \textbf{ticks} (the minimum price increment, typically \$0.01) or \textbf{basis points} (bps; 0.01\% of the price). They may also submit \textbf{market orders} for immediate execution against resting limit orders starting at the best bid/ask, and \textbf{cancellations} to withdraw resting orders. When an incoming order is matched with a resting one, a \textbf{fill} (or trade execution) occurs. This matching process is generally governed by a deterministic \textbf{price-time priority} algorithm, where orders are first prioritized by price and then by time of submission. These elements and mechanisms constitute \textbf{market microstructure}.

\subsection{Stylized Facts as Emergent Properties}\label{sec:background/stylized_facts}
The strategic interactions of market participants give rise to endogenous market dynamics~\citep{bouchaud2010endogenous}. These dynamics, in turn, give rise to universal and persistent statistical properties known as \textbf{stylized facts}. These facts are observed across a wide range of assets, markets, and time periods, and serve as a crucial benchmark for the realism of any generative market model~\citep{cont2001empirical, ratliff2023revisiting}. Key stylized facts include:
\begin{itemize}[leftmargin=20pt]
    \item \textbf{Heavy-Tailed Returns}: Returns are leptokurtic -- extreme movements occur far more frequently than a Gaussian predicts.
    \item \textbf{Volatility Clustering}: High-volatility periods cluster together, manifesting as slowly decaying autocorrelation of absolute returns.
    \item \textbf{Lack of Autocorrelation in Returns}: Consistent with efficient markets, return autocorrelation is insignificant beyond very short lags.
\end{itemize}

\section{Related Work} \label{sec:related_work}

The modeling of market microstructure has evolved from explicit, theory-driven formulations toward implicit, data-driven representation learning. Our work continues this trajectory, positioning a generative foundation model as the natural next step to learn universal market dynamics directly from raw, heterogeneous data.

\subsection{Market Microstructure and Order Flow Modeling}

\paragraph{Classical Stochastic Models}
A significant body of literature models order arrival times using point processes, such as Hawkes processes, to capture the self-exciting nature of order flow~\citep{bacry2015hawkes}. More sophisticated approaches adopt Compound Hawkes processes, which combine Hawkes processes to model interarrival times with other fitted empirical distributions to model additional features like volumes and price depths~\citep{jain_hawkes}. While providing strong theoretical grounding, these models rely on specific parametric assumptions (e.g., Gaussianity) that are unable to capture the heavy-tailed nature of market returns.
Similarly, ensemble methods based on Hidden Markov Models have been applied to classify sequences of trade activity, demonstrating that probabilistic sequence models capture meaningful structure in financial data even with limited labeled examples~\citep{kawawa2024ensemble}.

\paragraph{Agent-Based Models}
Agent-based models (ABMs) simulate market dynamics by defining the behavior of individual participants and observing the emergent properties of the system~\citep{abides}. While ABMs have historically required hand-crafting agent behaviors, recent approaches have shown success in calibrating agents on real market data~\citep{abides_economist}. Our work contributes to this line of research by enabling the learning of complex market dynamics, which can serve as a foundation for more sophisticated agent-based modeling.

\paragraph{Early Deep Learning Models}
The application of deep learning to LOB data was pioneered by models like DeepLOB~\citep{zhang2019deeplob}. These models demonstrated the potential of learning features directly from data but were typically trained on a subset of instruments. This limits their ability to learn universal representations across diverse assets and market conditions.

\subsection{Transformers and Foundation Models in Finance}
The Transformer architecture~\citep{vaswani2017attention} has been widely applied across domains including genomics~\citep{ji2021dnabert}, time-series forecasting~\citep{wen2022transformers}, and payment transaction modeling~\citep{raman2024scalable}.
In market microstructure, recent transformer models focus on discriminative prediction for short-term forecasting~\citep{berti2025tlob, xiao2025lit}. These approaches operate on full limit order book snapshots and target single or few instruments.
Generative adversarial networks~\citep{wiese2020quant} and diffusion-based approaches have also been applied to financial data generation. These methods typically target price or return series rather than event-level order flow, and have not been evaluated at multi-asset scale.

A prominent generative approach is MaRS~\citep{mars}, a market simulator with a foundation model backbone. While building on design principles from~\citet{mars}, TradeFM distinguishes itself in three dimensions (Table~\ref{tab:related_work_comparison}): (1)~pre-training data explicitly constructed to \textbf{maximize heterogeneity} across thousands of assets, sectors, and liquidity regimes; (2)~\textbf{partial observability} – learning from Level~3 trade messages rather than full LOB snapshots, matching the information available to real market participants; and (3)~\textbf{zero-shot geographic generalization} to unseen markets, enabled by scale-invariant feature design rooted in Universal Price Formation theory~\citep{sirignano2021universal}.
Direct quantitative comparison with MaRS is infeasible: it requires limit order book context as input (strictly more information than event streams), is trained on a different market (Chinese equities), and covers fewer assets (${\sim}$500 vs.\ $>$9K).

\begin{table}[htbp]
\centering
\small
\begin{tabular}{l|cccc}
\toprule
\textbf{Model} & \textbf{Input} & \textbf{Assets} & \textbf{Params} & \textbf{Zero-Shot} \\
\midrule
DeepLOB & Full LOB & 5 & 60K & No \\
LOBS5\nocite{nagy2023generative} & Full LOB & 2 & 6.3M & No \\
MaRS & Full LOB & 500 & $\sim$1B & No \\
\textbf{TradeFM} & \textbf{Event stream} & \textbf{$>$9K} & \textbf{524M} & \textbf{Yes} \\
\bottomrule
\end{tabular}
\caption{\textbf{Comparison with related microstructure models.} TradeFM uniquely combines partial observability, large-scale multi-asset training, and zero-shot geographic generalization.}
\label{tab:related_work_comparison}
\end{table}
\begin{figure*}[htbp]
    \centering
    \includegraphics[width=0.95\linewidth]{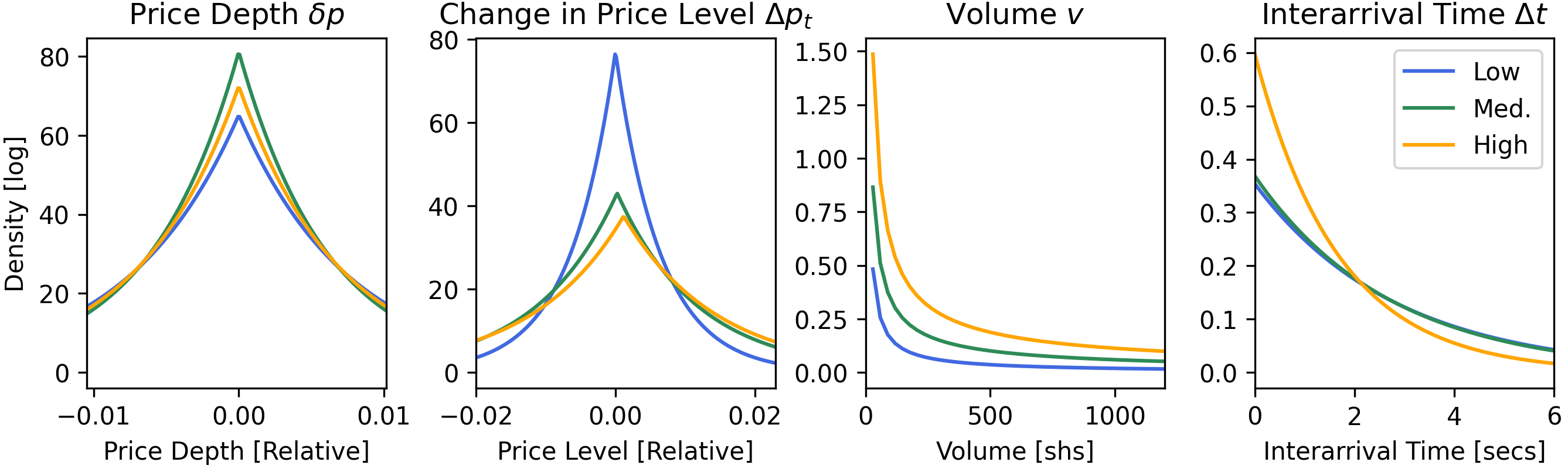}
    \vspace{-4pt}
    \caption{\textbf{Trade feature distributions.} Canonical distributions for core trade features, conditioned on liquidity. Price features are leptokurtic (Laplace); volume follows a heavy-tailed power-law; and interarrival time is exponential.}
    \label{fig:trade_data}
    \vspace{-16pt}
\end{figure*}

\section{Problem Formulation}
We formulate the task of modeling market microstructure as a generative, autoregressive sequence modeling problem. Let the market dynamics be represented by a sequence of discrete trade events, $E=(e_1, e_2, \ldots, e_T)$. The objective is to learn the conditional probability distribution $P(e_t | e_{<t})$, where $e_{<t}$ denotes the sequence of events preceding $e_t$. By learning this distribution, the model can generate realistic sequences of future trade events, simulating the market's evolution.

\subsection{Trade Event Representation}
A single trade event $e_t$ is a multi-feature tuple capturing the state of the market at the moment of a transaction. Formally, an event is represented as $e_t = (\Delta t_t, \delta p_t, v_t, a_t, s_t)$, where the core features are: $\Delta t_t$: Interarrival time since the previous event (seconds); $\delta p_t$: Price depth of the transaction (basis points); $v_t$: Volume of the transaction (shares); $a_t$: The action/order type (e.g., order submission, cancellation); $s_t$: The side of the initiating order (buy/bid or sell/ask). The distributions of these features are depicted in Figure~\ref{fig:trade_data}.

\subsection{Key Technical Challenges}
Modeling this data stream presents challenges inherent to high-frequency markets: the \textbf{Heterogeneity and Distribution Shift} across diverse assets and varying time periods; the \textbf{Sparsity and Irregularity} of the asynchronous event stream; the \textbf{Partial Observability} of the true market state from transaction data; and a \textbf{High-Dimensional, Multi-Modal Feature Space} combining continuous and categorical values.

\section{Data and Feature Engineering} \label{sec:data_pipeline}
Our methodology is designed to process raw, heterogeneous transaction data at scale and transform it into a standardized format suitable for a generative foundation model. This pipeline consists of data curation, robust feature engineering, and a novel tokenization scheme.

\subsection{Data Sources and Scale} \label{sec:data_split}
The model is pre-trained on a proprietary dataset built from billions of raw, tick-level US equities transactions, spanning 368 trading days from February 2024 to September 2025, across the breadth of the US equities market. This represents over 19 billion tokens across 1.9 million date-asset pairs.
We employ a temporal hold-out strategy, reserving January 2025 onward across all assets for the test set, yielding a training set of 10.7 billion tokens and a test set of 8.7 billion tokens. The tokenizer is calibrated on the first 30 days of the training data, February 2024. For evaluating out-of-distribution generalization we also hold out one month of data from APAC regions, namely Jan. 2025, for both Japan and China.

\subsection{Mid-Price Estimation}
A robust estimate of the true market mid-price ($\midprice$) is critical for normalizing price-related features.
In our partial-information setting, the mid-price is unobserved; we only see noisy execution prices ($\execprice$).
Building on the classical Volume-Weighted Average Price~\citep{berkowitz1988total}, we introduce \textbf{Exponentially-Weighted VWAP (EW-VWAP)}, a time-aware estimator that jointly accounts for trade volume and recency: $\hat{p}_t^{\text{EW-VWAP}} = \text{EMA}(\execprice \cdot v_t) / \text{EMA}(v_t)$.
The smoothing factor $\alpha$ is determined by a time-based halflife, ensuring that the estimate gives more weight to larger and more recent trades, providing a stable and representative price benchmark.
Unlike naive rolling averages, EW-VWAP is comparable across assets with vastly different liquidity profiles (see Appendix~\ref{sec:appendix/midprice_estimation}, Figure~\ref{fig:midprice_estimators}).

\subsection{Scale-Invariant Feature Construction} \label{sec:feature_construction}
The statistical properties of trading data vary widely across sectors, liquidity profiles, and nominal prices. In raw dollar, share, and second terms, price depths, volumes, and interarrival times for an asset like AAPL may differ greatly from those of a penny stock. Trade representations must therefore be carefully designed to enforce homogeneity across assets.
\citet{sirignano2021universal} demonstrate that price formation follows
universal principles across stocks: universal models trained on all assets outperform asset-specific models, even for held-out stocks, suggesting deep learning can learn appropriate normalizations from raw data. Building on this, we explicitly construct scale-invariant features to enforce homogeneity, extending universality from order book contexts to heterogeneous event streams across diverse trading venues and market structures.
\begin{itemize}[leftmargin=20pt]
\item \textbf{Interarrival Time ($\Delta t_t$)}: Wall clock time since the previous event: $w_t - w_{t-1}$, in seconds.

\item \textbf{Log-Transformed Volume ($v_t$)}: To compress the wide dynamic range of order sizes, which follow heavy-tailed, power-law distributions~\citep{vyetrenko2019realrealismmetricsrobust}, we apply a logarithmic transformation: $v_t = \log(1+V_t)$ where $V_t$ is the raw share volume.

\item \textbf{Normalized Price Depth ($d_t$)}: The depth of a limit order with order price $\orderprice$, relative to the mid-price: $d_t = \frac{\orderprice - \midpriceest}{\midpriceest}$, normalizing the raw $\delta p_t$ from the event tuple. This representation is comparable across differently priced assets, unlike prior work using price depths in ticks.

\item \textbf{Relative Price Level vs. Open ($\Delta p_t$)}: To capture intraday market movement, we quote the current mid-price relative to the day's opening price ($p_0$): $\Delta p_t = \frac{\midpriceest - p_0}{p_0}$.
\end{itemize}

While price-related features are computed as unit-less ratios, we refer to them in terms of basis points (bps) for interpretability, where a ratio of 0.01 corresponds to 100 bps.
Figure~\ref{fig:tick_vs_relative} (Appendix~\ref{sec:appendix/appendix_a}) demonstrates distributional stationarity of relative features (vs. tick based). Figure~\ref{fig:tokenizer_drift} (Appendix~\ref{sec:temporal_drift}) demonstrates the subsequent temporal stability of scale-invariant features.

\subsection{Market and Participant-Level Sequences}
In downstream applications, TradeFM can be used to model the behavior of the entire market (e.g., for synthetic data generation) or that of individual participants (e.g., for agent-based modeling). To support this, our training data includes sequences aggregated at both the market level and the participant level, with approximately a 1.6:1 market-to-participant ratio at the token level. The model receives a binary indicator feature, $I_{MP}$, to distinguish between these two contexts.

\section{Tokenization} \label{sec:tokenization}
Standard Transformer architectures, as applied in natural language processing, operate on univariate sequences where each element is a single token from a discrete vocabulary. Our trade event data is a sequence of multi-feature tuples, each comprising a mix of continuous and categorical values. The core challenge of tokenization is to map this event stream into a univariate discrete sequence.

\subsection{Binning Strategy and Outlier Handling}
\begin{figure}[htbp]
  \centering
    \includegraphics[width=0.55\textwidth]{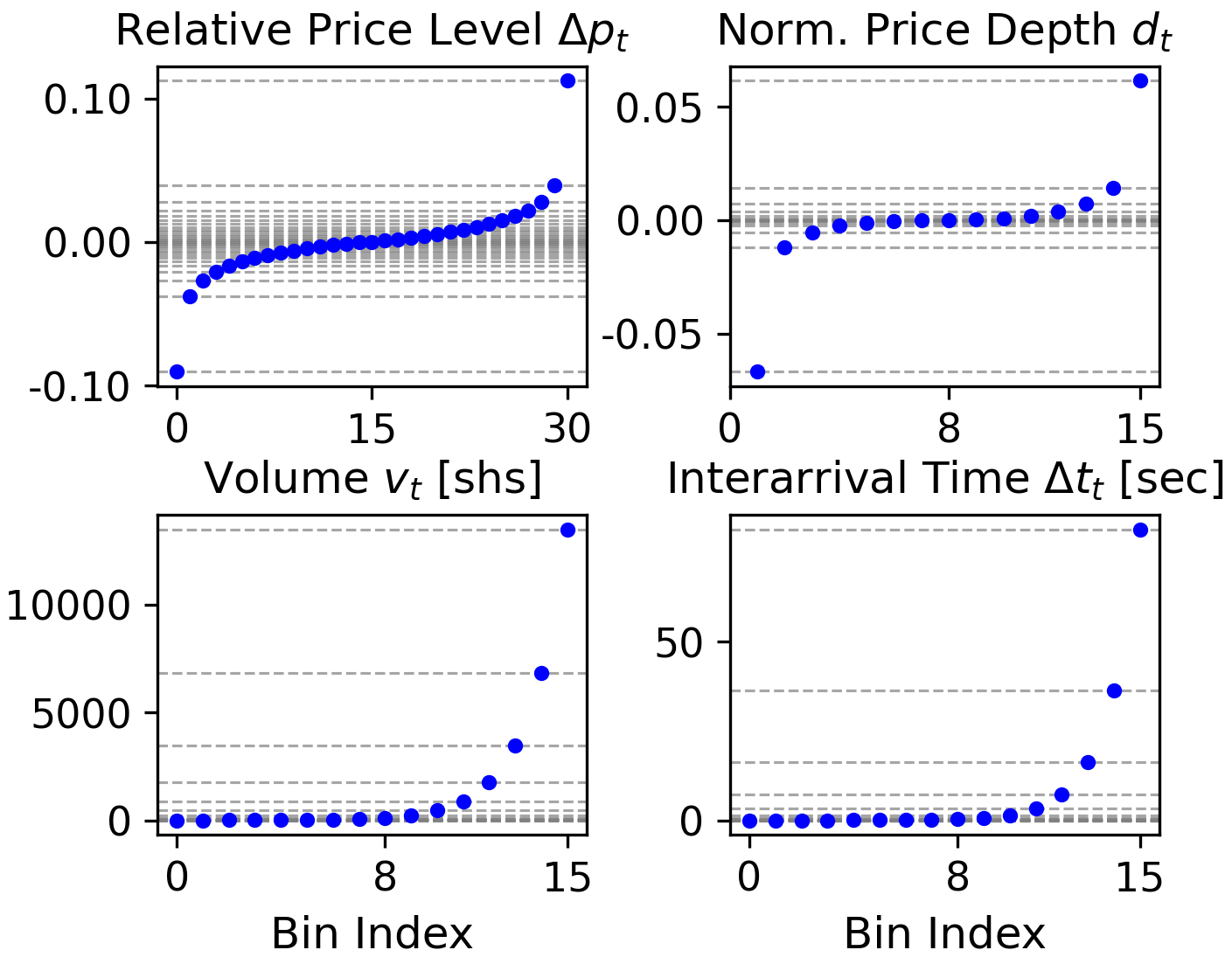}
    \caption{\textbf{Calibrated bin edges.} Price features (top) use quantile-based binning for high resolution near the mean; volume and time (bottom) use logarithmic bins to capture their wide dynamic range.}
    \label{fig:bin_edges}
    \vspace{-12pt}
\end{figure}
We discretize each continuous feature by partitioning its distribution into a fixed number of bins.
For price-related features, which have symmetric but highly peaked distributions, we employ \textbf{Equal-Frequency Binning} (quantile-based). This ensures that the bins in the dense central region of the distribution have a higher resolution, while still capturing the less frequent values in the tails.
For log-transformed features, like volume and interarrival time, we use \textbf{Equal-Width Binning}. Applying equal-width bins to the logarithmic values creates bins that are effectively logarithmic in the original feature space, providing a way to represent values that span multiple orders of magnitude.

This hybrid approach ensures a relatively uniform token distribution, preventing the model from wasting capacity on rare tokens or nearly-empty bins. Before binning, we exclude outliers above the 99th percentile for each feature, and additionally exclude outliers below the 1st percentile for price depth and price level. We reserve special bins to represent these out-of-range values to prevent the model from allocating excessive capacity to extremely rare events. We calibrate this tokenizer on the first 30 days of our data. The calibrated bin edges are shown in Figure~\ref{fig:bin_edges}.

\subsection{Multi-Feature Token Composition}\label{sec:multi_feature_token}
While our model's input at each time step is multi-featured, the decoder is trained to predict a single, unidimensional token, representing the core trade event. Thus, we combine the discrete bin indices of the trade-related features, ($i_{\Delta t}, i_{\delta p}, i_v, i_a, i_s$), into a single composite integer, $i_{\text{trade}}$.

This is accomplished by treating the feature indices as digits in a mixed base number system. Each feature's bin index is a ``digit'', and the number of possible values for the subsequent features acts as the ``base'' at each position. For a concrete example of the encoding process, see Section~\ref{sec:sample_encoding}. 
The number of bins for each feature is: $n_{\delta p}=16$ for price depth, $n_v=16$ for volume, $n_{\Delta t}=16$ for interarrival time, $n_s=2$ (buy, sell) for side, and $n_a=2$ (add or cancel order) for action type. The composite trade token $i_{\text{trade}}$, a single integer encoding all constituent features, is calculated as:
\begin{equation}
\begin{aligned}
    i_{\text{trade}} = & (i_a \times n_s \times n_{\delta p} \times n_v \times n_{\Delta t}) + \\ &(i_s \times n_{\delta p} \times n_v \times n_{\Delta t}) + \\ 
    & (i_{\delta p} \times n_v \times n_{\Delta t}) + (i_{v} \times n_{\Delta t}) + i_{\Delta t}
\end{aligned}
\label{eq:tokenization}
\end{equation}

This yields a vocabulary size of 16,384 for the predictable trade tokens.
The model input at each time step is a tuple containing this token along with several non-predicted features used for conditioning. These contextual features are provided as separate inputs, and are not part of the trade token $i_{\text{trade}}$ calculated in Equation~\ref{eq:tokenization}.
The contextual features are:
\begin{itemize}[leftmargin=20pt]
    \item $i_l$: The liquidity bin index ($n_l=3$), determined by binning each asset into low, medium, or high liquidity ranges based on its Average Daily Volume (ADV).
    \item $i_{\Delta p_t}$: The price level change bin index ($n_{\Delta p}=32$).
    \item $I_{MP}$: A binary indicator for market-level vs. participant-level sequences.
\end{itemize}
The final input is $[i_l, I_{MP}, i_{\Delta p_t}, i_{\text{trade}}]$. This formulation allows the model to be conditioned on broader market context while focusing its predictive power on the next trade event.

\section{TradeFM Architecture}\label{sec:TradeFM}
\textbf{TradeFM is a decoder-only Transformer}, trained from scratch with a custom configuration. The architecture is based on the Llama family and incorporates enhancements including grouped-query-attention (GQA) and rotary positional encoding (RoPE) \citep{llama}. Our model size is \textbf{524 million parameters}, a size chosen based on Chinchilla scaling laws \citep{chinchilla_scaling} for our dataset size (see Appendix~\ref{sec:TradeFM/Hyperparams} for detailed hyperparameter choices). The model is trained on \textbf{3 Nvidia A100 GPUs}; we include detailed training setup details in Appendix~\ref{sec:TradeFM/Training}.

\subsection{Tabular Input Embedding}\label{sec:TradeFM/InputLayer}
We employ a tabular embedding approach to handle our multi-feature input tokens (as described in Section~\ref{sec:multi_feature_token}). Each feature in the input tuple $[i_l, I_{MP}, i_{\Delta p_t}, i_{\text{trade}}]$ is first projected into its own embedding space using an embedding table. These embedding vectors are concatenated and passed through a linear projection layer to create a unified representation in the model's hidden dimension.
\section{Market Simulator} \label{sec:market_simulator}
To evaluate the realism of our generative model, we require an environment that can simulate the market execution of a sequence of predicted trades. We build a lightweight, deterministic simulator tailored to our specific experimental setting. Our simulator serves two critical functions: 1) it provides the dynamic, state-dependent price level features required by our model during generative rollouts, and 2) it allows us to test if the model's generated trade flow can reproduce the well-known stylized facts of asset returns.

\begin{figure}[htbp]
    \centering
    \includegraphics[width=0.5\textwidth]{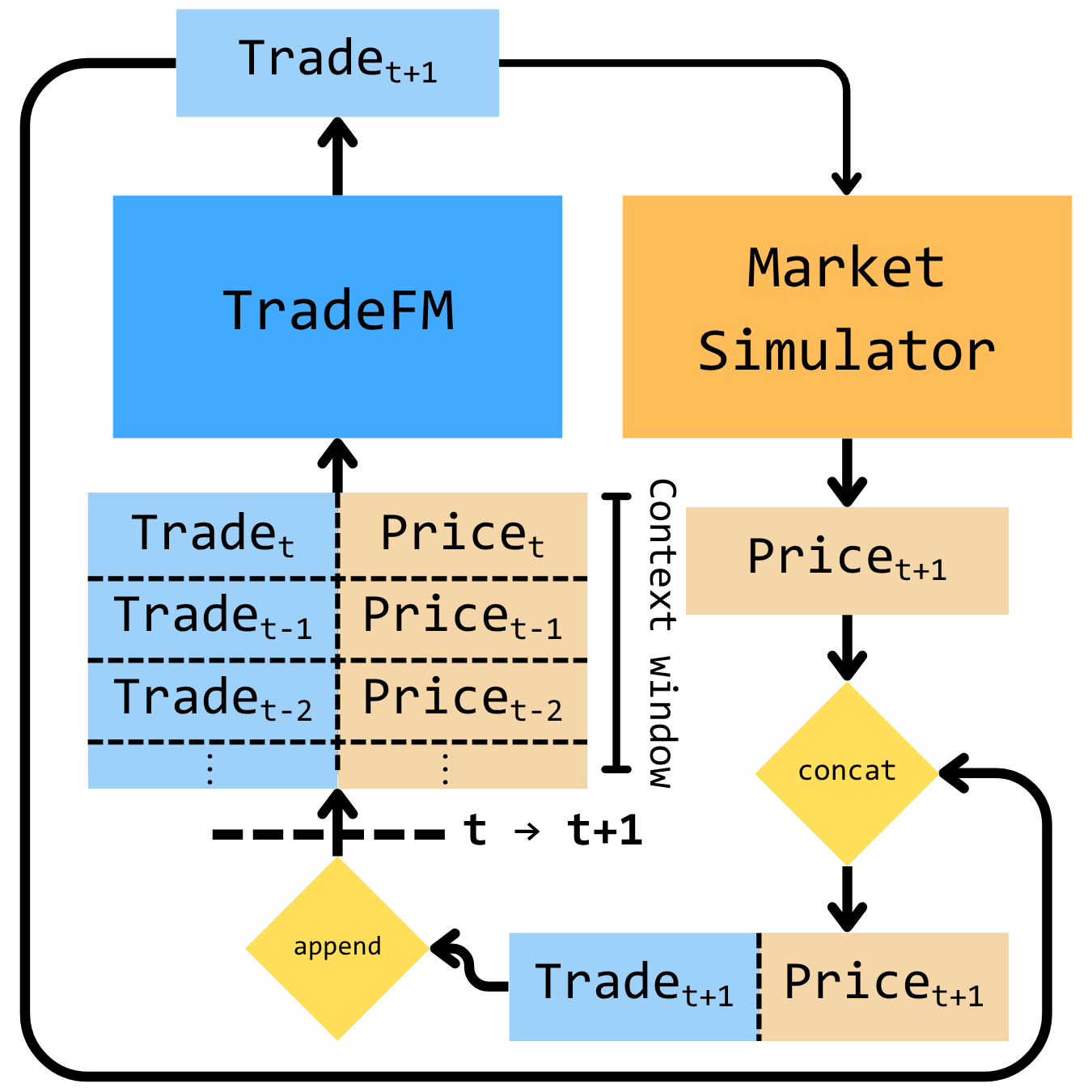}
    \caption{\textbf{Closed-loop simulation architecture.} TradeFM predicts a trade, the Market Simulator executes it, and the updated market state is fed back to the model.}
    \label{fig:sim_model_diagram}
\end{figure}

\subsection{Deterministic Design}
The simulator is designed to mimic core mechanics of a modern electronic exchange. It maintains a limit order book (LOB) for an asset, an internal clock, and an estimate of the market mid-price (the midpoint of the best bid and ask). The order matching engine employs \textbf{price-time priority}: incoming orders are matched with the best-priced order on the opposite side of the book, with ties in price broken by selecting the earliest-submitted resting order~\citep{nasdaq}. Before using the simulator to validate our large trade model, \textbf{we validate the realism of the simulator itself via stylized facts} discussed in the Appendix~\ref{sec:sim_validation}.
\subsection{The Closed-Loop Rollout}
The simulator creates a closed-loop system where the model and environment interact dynamically. This process, which we term a \textit{rollout}, is shown in Figure~\ref{fig:sim_model_diagram} and proceeds as follows:
\begin{enumerate}[leftmargin=20pt]
    \item \textbf{Prediction}: Given a history of market events, TradeFM predicts the next event token, $i_{\text{trade}}$. We use multinomial sampling with a repetition penalty of 1.2 to decode the token.
    \item \textbf{Execution}: The predicted event (e.g., order or cancellation) is passed to the simulator, which executes it against the LOB according to its price-time priority rules.
    \item \textbf{State Update}: The simulator updates its internal state, including the LOB and the mid-price.
    \item \textbf{Feedback}: The market state is used to generate the contextual features for the next time step, which are appended to the history and fed back into TradeFM to generate the next prediction.
\end{enumerate}
This recursive loop allows us to generate long, dynamic sequences of market activity. Crucially, it enables the study of second-order effects like price impact, as the model's own predictions influence the market state that conditions its future predictions.

\section{Experiments}
Evaluating generative models of financial markets presents a unique challenge due to the non-stationary nature of the data. 
Rather than relying on next-token perplexity, we evaluate TradeFM by its ability to reproduce the invariant stylized facts of market behavior (Section~\ref{sec:background/stylized_facts}).
A model that generates synthetic data exhibiting these facts has learned the time-invariant structure of the market, not merely memorized historical patterns.
We conduct three experiments: (1)~\textbf{stylized fact reproduction}, (2)~\textbf{distributional fidelity} of generated order flow, and (3)~out-of-distribution \textbf{generalization and controllability}.
\paragraph{Baselines} 
We compare against a \textbf{calibrated Zero-Intelligence (ZI) agent}~\citep{zi_traders, farmer2005predictive} and a \textbf{Compound Hawkes} process~\citep{bacry2015hawkes, jain_hawkes} that models trade arrivals via self-exciting dynamics with fitted volume and price distributions. Both baselines interact with the same simulator and evaluation pipeline. Details in Appendix~\ref{sec:appendix/zi_baseline}.
We adopt ZI and Compound Hawkes as baselines because they operate on trade-level event streams, matching our partial observability setting. Neural generative models evaluated in LOB-Bench~\citep{nagy2025lob} require full limit order book input, precluding direct comparison.
\subsection{Experiment 1: Stylized Fact Reproduction}
\label{sec:exp_1}
To validate the realism of the generated market trajectories, we evaluate their ability to reproduce key stylized facts of log returns ($r_{t,\Delta t}$). We generate 10 rollouts of 1,024 events for 9 assets across 3 liquidity tiers, for each of 9 held-out months, conditioned on a context of 1,024 real historical events. 
For reference, 1,024 events correspond to approximately 2--5 minutes of market time for high-liquidity assets (e.g., AAPL), 15--60 minutes for medium-liquidity, and 1--4 hours for low-liquidity assets.
We compute autocorrelations over time lags $\tau$, and evaluate kurtosis over return intervals $\Delta t_r$.

\begin{table*}[htbp]
\centering
\begin{tabular}{c|ccc|ccc}
\toprule
\textbf{$\Delta t_r$} & \multicolumn{3}{c|}{\textbf{Kolmogorov-Smirnov}} & \multicolumn{3}{c}{\textbf{Wasserstein}} \\
 & \textbf{ZI} & \textbf{Hawkes} & \textbf{TradeFM} & \textbf{ZI} & \textbf{Hawkes} & \textbf{TradeFM} \\
\midrule
10  & 0.376 & 0.039 & \textbf{0.013} & 0.0006 & 0.0004 & \textbf{0.0001} \\
30  & 0.439 & 0.075 & \textbf{0.024} & 0.0015 & 0.0012 & \textbf{0.0002} \\
60  & 0.429 & 0.101 & \textbf{0.035} & 0.0027 & 0.0023 & \textbf{0.0004} \\
120 & 0.429 & 0.098 & \textbf{0.043} & 0.0043 & 0.0043 & \textbf{0.0007} \\
\bottomrule
\end{tabular}
\caption{\textbf{Log return distributional fidelity.} Mean K-S and Wasserstein distances from real data across return intervals $\Delta t_r$ (sec), averaged over 9 assets, 3 liquidity tiers, and 9 held-out months. TradeFM achieves 2--3$\times$ lower K-S distance than Compound Hawkes.}
\label{tab:log_return_distances}
\vspace{-8pt}
\end{table*}

\begin{figure*}[htbp]
    \centering
    \includegraphics[width=0.95\linewidth]{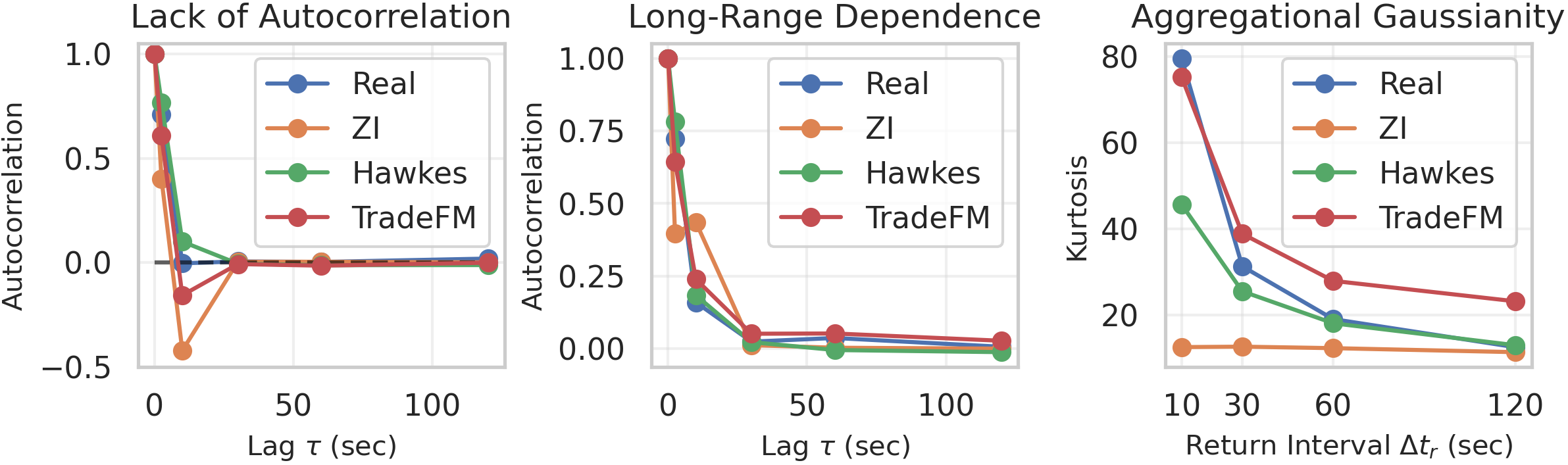}
    \caption{\textbf{TradeFM model validation.} Simulated returns exhibit: (Left) near-zero autocorrelation, (Middle) slowly decaying autocorrelation of absolute returns (volatility clustering), and (Right) heavy tails and aggregational Gaussianity.}
    \label{fig:log_return_stylized_facts}
    \vspace{-12pt}
\end{figure*}

\paragraph{Results} As summarized in Figure~\ref{fig:log_return_stylized_facts} and Table~\ref{tab:log_return_distances}, our simulations successfully reproduce the target stylized facts, demonstrating a close correspondence with real market data:
\begin{itemize}[leftmargin=20pt]
    \item \textbf{Lack of Autocorrelation}: The autocorrelation of simulated log returns (left panel) quickly decays to statistically insignificant levels as the lag $\tau$ increases. This is consistent with the efficient market hypothesis. The ZI baseline exhibits spurious autocorrelation.
    \item \textbf{Long-Range Dependence}: The autocorrelation of absolute log returns (middle panel) decays slowly, indicating that our model has captured the long-memory nature of volatility clustering.
    \item \textbf{Heavy Tails}: The kurtosis of simulated returns (right) is high for short time scales ($\Delta t_r$), confirming the presence of heavy tails. TradeFM most faithfully captures the leptokurtic nature of returns and its decay across time scales.
    \item \textbf{Aggregational Gaussianity}: As $\Delta t_r$ increases, the kurtosis of TradeFM correctly approaches that of a normal distribution, capturing the reversion towards normality over longer time horizons.
\end{itemize}

Table~\ref{tab:log_return_distances} quantifies these results via the Wasserstein distance ($W_1$) and Kolmogorov-Smirnov (K-S) statistic between real and generated log return distributions. TradeFM outperforms all baselines in both metrics across return intervals.

\subsection{Experiment 2: Quantitative Fidelity}

While stylized facts confirm that the model captures emergent market dynamics, they do not assess how well the generated order flow aligns with reality. We conduct a quantitative evaluation of distributional fidelity, adopting frameworks established in recent benchmarks for generative order flow models \citep{nagy2025lob}. As in the benchmark, we mean-variance normalize distributions before computing Wasserstein distance to make this metric comparable between quantities.

Using the rollouts described in Section~\ref{sec:exp_1}, we compute the Kolmogorov-Smirnov statistic and Wasserstein distance between real and generated distributions for key microstructure variables, including Order Volume, Interarrival Times, Bid-Ask Spreads, and Order Book Imbalance.
This evaluation averages results across 9 assets, 3 liquidity tiers, and 9 held-out months. As shown in Table~\ref{tab:order_flow_distances}, TradeFM achieves lower distance metrics than baseline approaches on most quantities, demonstrating superior fidelity in reproducing the statistical properties of market data.
The notable exception is bid-ask spreads, where the Compound Hawkes baseline achieves lower K-S distance.
We attribute this to the Hawkes process explicitly modeling inter-arrival dynamics that directly govern spread formation, whereas TradeFM's spread emerges indirectly from the simulator's deterministic order matching logic.
This suggests that hybrid approaches combining learned order flow with explicit spread modeling may yield further improvements.
We provide detailed results in Appendix~\ref{sec:temporal_drift}, Figure~\ref{fig:obi_over_time}.

The closed-loop evaluation couples model quality with simulator fidelity. Two lines of evidence partially disentangle their contributions: the ZI control through the identical simulator fails to reproduce stylized facts, and Table~\ref{tab:order_flow_distances} evaluates order flow directly before simulator interaction. Full disentanglement remains future work.

\begin{table*}[htbp]
\centering
\begin{tabular}{l|ccc|ccc}
\toprule
\multirow{2}{*}{\textbf{Quantity}} & \multicolumn{3}{c|}{\textbf{Kolmogorov-Smirnov}} & \multicolumn{3}{c}{\textbf{Wasserstein}} \\
 & \textbf{ZI} & \textbf{Hawkes} & \textbf{TradeFM} & \textbf{ZI} & \textbf{Hawkes} & \textbf{TradeFM} \\
\midrule
Spreads & 0.400 & \textbf{0.218} & 0.238 & 0.375 & \textbf{0.302} & 0.400 \\
Interarrival Times   & 0.651 & 0.515 & \textbf{0.281} & 0.415 & 0.626 & \textbf{0.318} \\
Price Depth & 0.436 & 0.281 & \textbf{0.169} & 0.390 & 0.348 & \textbf{0.339} \\
Order Book Imbalance  & 0.237 & 0.155 & \textbf{0.142} & 0.200 & 0.165 & \textbf{0.099} \\
Bid Volume          & 0.460 & \textbf{0.296} & 0.386 & 0.616 & 0.278 & \textbf{0.130} \\
Ask Volume         & 0.391 & 0.380 & \textbf{0.360} & 0.638 & 0.198 & \textbf{0.160} \\
\bottomrule
\end{tabular}
\caption{\textbf{Distributional fidelity of generated order flow.} Mean K-S and Wasserstein distances across 9 assets, 3 liquidity tiers, and 9 held-out months. TradeFM achieves lowest distance on most quantities. Per-month breakdowns in Appendix~\ref{sec:temporal_drift}, Figure~\ref{fig:obi_over_time}.}
\label{tab:order_flow_distances}
\vspace{-12pt}
\end{table*}

\subsection{Experiment 3: Generalization and Controllability}

To validate our claim that TradeFM learns a universal grammar of market microstructure that generalizes beyond the assets and time periods seen during training, we perform extensive out-of-distribution (OOD) evaluations.

\paragraph{Temporal and Geographic Robustness} Financial markets are non-stationary, with regimes shifting over time. We evaluate model performance over a hold-out period from January--September 2025, a period exhibiting heightened volatility distinct from the 2024 training set. As detailed in Appendix~\ref{sec:temporal_drift}, Figure~\ref{fig:obi_over_time}, distance metrics remain stable over this 9-month horizon.
\begin{figure}[htbp]
    \centering
    \includegraphics[width=0.53\textwidth]{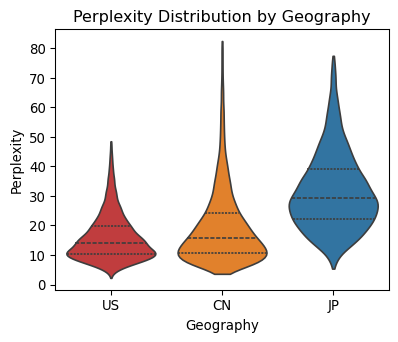}
    \caption{\textbf{Zero-shot geographic generalization.} Perplexity distributions for TradeFM (trained on US equities) evaluated on held-out US, China, and Japan data (January 2025) demonstrate cross-geography robustness of scale-invariant features.}
    \label{fig:geographic_ood}
\end{figure}

Beyond temporal generalization, we assess zero-shot geographic transfer by evaluating TradeFM (trained exclusively on US equities) on held-out data from China and Japan. Despite substantial microstructural differences – including Japan's Itayose batch auction mechanism (vs.\ US continuous trading), China's $\pm 10\%$ daily price limits (vs.\ none in US), and wider bid-ask spreads in Asian markets (3--10 bps vs.\ 1--2 bps) – Figure~\ref{fig:geographic_ood} shows moderate perplexity degradation, with significant overlap between US and APAC distributions. This cross-market generalization demonstrates that scale-invariant order flow representations capture universal principles of market dynamics.

\paragraph{Controllability}
Finally, we verify that the model respects its conditioning tokens and that its output can be reliably steered via the indicator features ($I_{MP}$ and $i_l$). We generate 256 context-free trajectories of 512 tokens each, for every combination of market-participant and liquidity indicators. We then analyze the statistical properties of the raw generated orders by computing the standard deviation of volumes and interarrival times for each condition.

Figure~\ref{fig:controllability_experiments} shows that modulating the indicator tokens has a significant and intuitive effect on the statistical properties of the generated orders, with two trends:
\begin{figure}[htbp]
    \centering
    \includegraphics[width=0.65\textwidth]{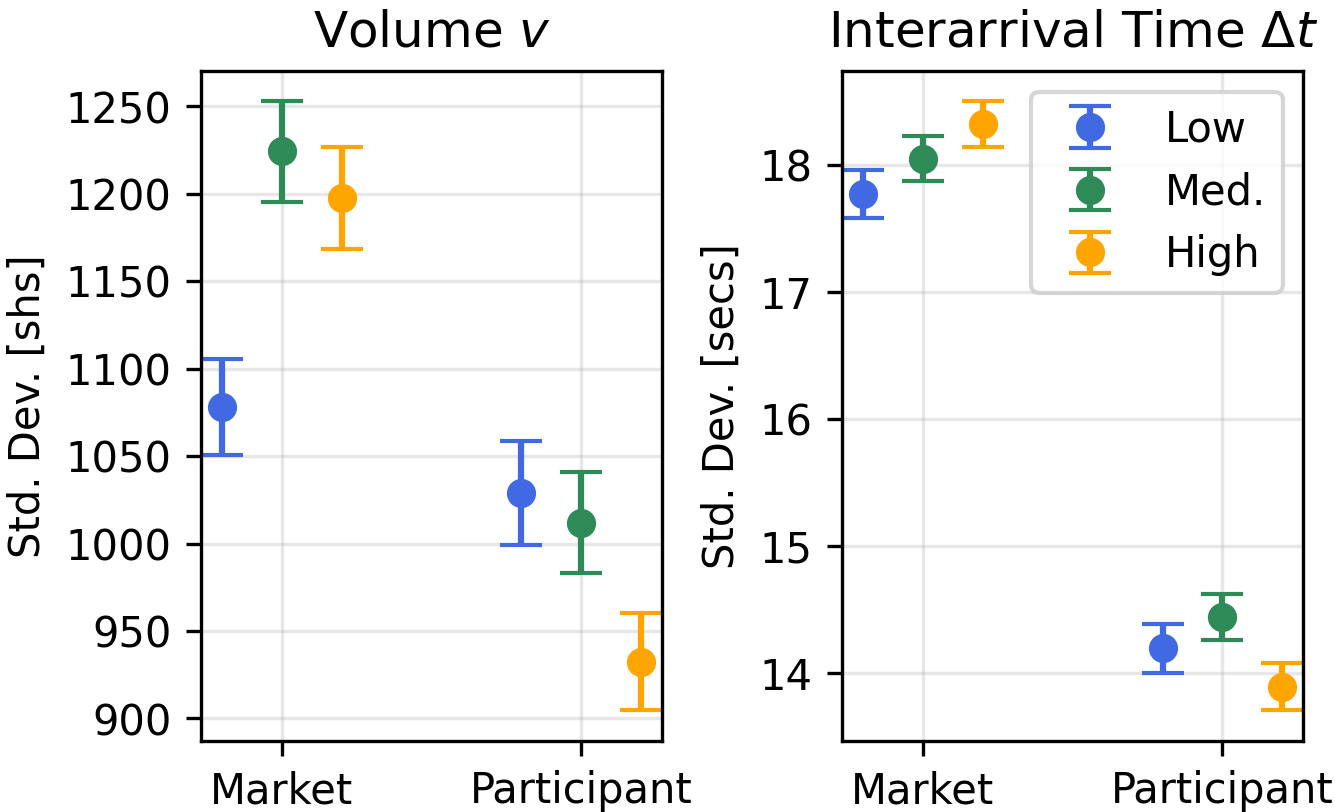}
    \caption{\textbf{Controllability experiments.} Standard deviation of generated volumes and interarrival times, conditioned on liquidity ($i_l$) and observation level ($I_{MP}$). The model produces distinct order flow, demonstrating controllable generation.}
    \label{fig:controllability_experiments}
    \vspace{-14pt}
\end{figure}

\begin{itemize}[leftmargin=20pt]
    \item The variance of both volume and interarrival time is consistently higher for market-level generation than for participant-level, aligned with the intuition that the aggregate behavior of an entire market is inherently more volatile than the behavior of a single participant.
    \item The model captures linear relationships between liquidity and the variance of interarrival times and volumes.
\end{itemize}
Collectively, these results demonstrate that TradeFM has learned a generalizable, conditional model of market behavior, capable of generating statistically and contextually appropriate order flow.

\section{Conclusion}
We have shown that the complex, emergent dynamics of financial markets can be learned directly from raw, heterogeneous order flow.
Our end-to-end methodology, which combines scale-invariant feature representations with a universal tokenization scheme, allows a single generative Transformer to generalize across thousands of diverse assets without asset-specific calibration.
The high fidelity of generated data – validated at both the emergent price dynamics level (Figure~\ref{fig:log_return_stylized_facts}) and the granular order level (Figure~\ref{fig:order_stylized_facts}, Appendix~\ref{sec:synthetic_data}) – combined with zero-shot geographic generalization suggests that scale-invariant trade representations capture fundamental, transferable structure in market microstructure.
The closed-loop simulator opens several directions for future work: privacy-preserving synthetic data generation for illiquid assets, stress testing under counterfactual scenarios, and training learning-based trading agents via interaction with the simulator (preliminary explorations in Appendix~\ref{sec:synthetic_data},~\ref{sec:market_sim_stress_test},~\ref{sec:multi_agent_modeling}).
Validating TradeFM's utility for these downstream applications remains a key priority.

\subsection*{Disclaimer}
This paper was prepared for informational purposes by the Artificial Intelligence Research group of JPMorgan Chase \& Co. and its affiliates ("JP Morgan'') and is not a product of the Research Department of JP Morgan. JP Morgan makes no representation and warranty whatsoever and disclaims all liability, for the completeness, accuracy or reliability of the information contained herein. This document is not intended as investment research or investment advice, or a recommendation, offer or solicitation for the purchase or sale of any security, financial instrument, financial product or service, or to be used in any way for evaluating the merits of participating in any transaction, and shall not constitute a solicitation under any jurisdiction or to any person, if such solicitation under such jurisdiction or to such person would be unlawful.

\bibliographystyle{icml2026}
\bibliography{references}

\begin{thebibliography}{37}
\providecommand{\natexlab}[1]{#1}
\providecommand{\url}[1]{\texttt{#1}}
\expandafter\ifx\csname urlstyle\endcsname\relax
  \providecommand{\doi}[1]{doi: #1}\else
  \providecommand{\doi}{doi: \begingroup \urlstyle{rm}\Url}\fi

\bibitem[Bacry et~al.(2015)Bacry, Mastromatteo, and Muzy]{bacry2015hawkes}
Bacry, E., Mastromatteo, I., and Muzy, J.-F.
\newblock {Hawkes} processes in finance.
\newblock \emph{Market Microstructure and Liquidity}, 1\penalty0 (01):\penalty0
  1550005, 2015.

\bibitem[Badaro et~al.(2023)Badaro, Saeed, and Papotti]{badaro2023transformers}
Badaro, G., Saeed, M., and Papotti, P.
\newblock Transformers for tabular data representation: A survey of models and
  applications.
\newblock \emph{Transactions of the Association for Computational Linguistics},
  11:\penalty0 227--249, 2023.

\bibitem[Berkowitz et~al.(1988)Berkowitz, Logue, and
  Noser~Jr.]{berkowitz1988total}
Berkowitz, S.~A., Logue, D.~E., and Noser~Jr., E.~A.
\newblock The total cost of transactions on the {NYSE}.
\newblock \emph{The Journal of Finance}, 43\penalty0 (1):\penalty0 97--112,
  1988.

\bibitem[Berti \& Kasneci(2025)Berti and Kasneci]{berti2025tlob}
Berti, L. and Kasneci, G.
\newblock {TLOB}: A novel transformer model with dual attention for stock price
  trend prediction with limit order book data.
\newblock \emph{arXiv preprint arXiv:2502.15757}, 2025.

\bibitem[Bouchaud(2010)]{bouchaud2010endogenous}
Bouchaud, J.-P.
\newblock The endogenous dynamics of markets: Price impact and feedback loops.
\newblock \emph{arXiv preprint arXiv:1009.2928}, 2010.

\bibitem[Byrd et~al.(2020)Byrd, Hybinette, and Balch]{abides}
Byrd, D., Hybinette, M., and Balch, T.~H.
\newblock {ABIDES}: Towards high-fidelity multi-agent market simulation.
\newblock In \emph{Proceedings of the 2020 ACM SIGSIM Conference on Principles
  of Advanced Discrete Simulation}, pp.\  11--22, 2020.

\bibitem[Cont(2001)]{cont2001empirical}
Cont, R.
\newblock Empirical properties of asset returns: Stylized facts and statistical
  issues.
\newblock \emph{Quantitative Finance}, 1\penalty0 (2):\penalty0 223--236, 2001.

\bibitem[Dwarakanath et~al.(2024)Dwarakanath, Vyetrenko, Tavallali, and
  Balch]{abides_economist}
Dwarakanath, K., Vyetrenko, S., Tavallali, P., and Balch, T.
\newblock {ABIDES-Economist}: Agent-based simulator of economic systems with
  learning agents.
\newblock \emph{arXiv preprint arXiv:2402.09563}, 2024.

\bibitem[Farmer et~al.(2005)Farmer, Patelli, and Zovko]{farmer2005predictive}
Farmer, J.~D., Patelli, P., and Zovko, I.~I.
\newblock The predictive power of zero intelligence in financial markets.
\newblock \emph{Proceedings of the National Academy of Sciences}, 102\penalty0
  (6):\penalty0 2254--2259, 2005.

\bibitem[Garza et~al.(2023)Garza, Challu, and
  Mergenthaler-Canseco]{garza2023timegpt}
Garza, A., Challu, C., and Mergenthaler-Canseco, M.
\newblock {TimeGPT}-1.
\newblock \emph{arXiv preprint arXiv:2310.03589}, 2023.

\bibitem[Gode \& Sunder(1993)Gode and Sunder]{zi_traders}
Gode, D.~K. and Sunder, S.
\newblock Allocative efficiency of markets with zero-intelligence traders:
  Market as a partial substitute for individual rationality.
\newblock \emph{Journal of Political Economy}, 101\penalty0 (1):\penalty0
  119--137, 1993.

\bibitem[Hasbrouck(2007)]{hasbrouck2007empirical}
Hasbrouck, J.
\newblock \emph{Empirical Market Microstructure: The Institutions, Economics,
  and Econometrics of Securities Trading}.
\newblock Oxford University Press, 2007.

\bibitem[Hoffmann et~al.(2022)Hoffmann, Borgeaud, Mensch, Buchatskaya, Cai,
  Rutherford, de~Las~Casas, Hendricks, Welbl, Clark, Hennigan, Noland,
  Millican, van~den Driessche, Damoc, Guy, Osindero, Simonyan, Elsen, Rae,
  Vinyals, and Sifre]{chinchilla_scaling}
Hoffmann, J., Borgeaud, S., Mensch, A., Buchatskaya, E., Cai, T., Rutherford,
  E., de~Las~Casas, D., Hendricks, L.~A., Welbl, J., Clark, A., Hennigan, T.,
  Noland, E., Millican, K., van~den Driessche, G., Damoc, B., Guy, A.,
  Osindero, S., Simonyan, K., Elsen, E., Rae, J.~W., Vinyals, O., and Sifre, L.
\newblock Training compute-optimal large language models.
\newblock In \emph{Advances in Neural Information Processing Systems},
  volume~35, pp.\  30016--30030, 2022.

\bibitem[Hollmann et~al.(2025)Hollmann, M{\"u}ller, Purucker, Krishnakumar,
  K{\"o}rfer, Hoo, Schirrmeister, and Hutter]{hollmann2025accurate}
Hollmann, N., M{\"u}ller, S., Purucker, L., Krishnakumar, A., K{\"o}rfer, M.,
  Hoo, S.~B., Schirrmeister, R.~T., and Hutter, F.
\newblock Accurate predictions on small data with a tabular foundation model.
\newblock \emph{Nature}, 637\penalty0 (8045):\penalty0 319--326, 2025.

\bibitem[Jain et~al.(2024)Jain, Firoozye, Kochems, and Treleaven]{jain_hawkes}
Jain, K., Firoozye, N., Kochems, J., and Treleaven, P.
\newblock Limit order book dynamics and order size modelling using compound
  {Hawkes} process.
\newblock \emph{Finance Research Letters}, 69:\penalty0 106157, 2024.

\bibitem[Ji et~al.(2021)Ji, Zhou, Liu, and Davuluri]{ji2021dnabert}
Ji, Y., Zhou, Z., Liu, H., and Davuluri, R.~V.
\newblock {DNABERT}: Pre-trained bidirectional encoder representations from
  transformers model for {DNA}-language in genome.
\newblock \emph{Bioinformatics}, 37\penalty0 (15):\penalty0 2112--2120, 2021.

\bibitem[Kaplan et~al.(2020)Kaplan, McCandlish, Henighan, Brown, Chess, Child,
  Gray, Radford, Wu, and Amodei]{kaplan_scaling}
Kaplan, J., McCandlish, S., Henighan, T., Brown, T.~B., Chess, B., Child, R.,
  Gray, S., Radford, A., Wu, J., and Amodei, D.
\newblock Scaling laws for neural language models.
\newblock \emph{arXiv preprint arXiv:2001.08361}, 2020.

\bibitem[Kawawa-Beaudan et~al.(2024)Kawawa-Beaudan, Sood, Palande, Mani, Balch,
  and Veloso]{kawawa2024ensemble}
Kawawa-Beaudan, M., Sood, S., Palande, S., Mani, G., Balch, T., and Veloso, M.
\newblock Ensemble methods for sequence classification with hidden {Markov}
  models.
\newblock \emph{arXiv preprint arXiv:2409.07619}, 2024.

\bibitem[Li et~al.(2025)Li, Liu, Liu, Fang, Wang, Xu, and Bian]{mars}
Li, J., Liu, Y., Liu, W., Fang, S., Wang, L., Xu, C., and Bian, J.
\newblock {MarS}: A financial market simulation engine powered by generative
  foundation model.
\newblock In \emph{Proceedings of the Thirteenth International Conference on
  Learning Representations}, 2025.

\bibitem[Muennighoff et~al.(2023)Muennighoff, Rush, Barak, Le~Scao, Piktus,
  Tazi, Pyysalo, Wolf, and Raffel]{scaling_data_constrained}
Muennighoff, N., Rush, A.~M., Barak, B., Le~Scao, T., Piktus, A., Tazi, N.,
  Pyysalo, S., Wolf, T., and Raffel, C.
\newblock Scaling data-constrained language models.
\newblock In \emph{Advances in Neural Information Processing Systems},
  volume~36, 2023.

\bibitem[Nagy et~al.(2023)Nagy, Frey, Sapora, Li, Calinescu, Zohren, and
  Foerster]{nagy2023generative}
Nagy, P., Frey, S., Sapora, S., Li, K., Calinescu, A., Zohren, S., and
  Foerster, J.
\newblock Generative {AI} for end-to-end limit order book modelling: A
  token-level autoregressive generative model of message flow using a deep
  state space network.
\newblock In \emph{Proceedings of the Fourth ACM International Conference on AI
  in Finance}, pp.\  91--99, 2023.

\bibitem[Nagy et~al.(2025)Nagy, Frey, Li, Sarkar, Vyetrenko, Zohren, Calinescu,
  and Foerster]{nagy2025lob}
Nagy, P., Frey, S., Li, K., Sarkar, B., Vyetrenko, S., Zohren, S., Calinescu,
  A., and Foerster, J.
\newblock {LOB-Bench}: Benchmarking generative {AI} for finance -- an
  application to limit order book data.
\newblock In \emph{Proceedings of the Forty-Second International Conference on
  Machine Learning}, 2025.

\bibitem[{Nasdaq Listing Center}(2024)]{nasdaq}
{Nasdaq Listing Center}.
\newblock Equity trading rules.
\newblock
  \url{https://listingcenter.nasdaq.com/rulebook/nasdaq/rules/Nasdaq\%20Equity\%204},
  2024.

\bibitem[Pasca(2015)]{Pasca2015151167}
Pasca, L.
\newblock A critical review of the main approaches on financial market dynamics
  modelling.
\newblock \emph{Journal of Heterodox Economics}, 2\penalty0 (2):\penalty0
  151--167, 2015.
\newblock \doi{10.1515/jheec-2015-0017}.

\bibitem[Petty et~al.(2024)Petty, van Steenkiste, Dasgupta, Sha, Garrette, and
  Linzen]{llm_depth_width}
Petty, J., van Steenkiste, S., Dasgupta, I., Sha, F., Garrette, D., and Linzen,
  T.
\newblock The impact of depth on compositional generalization in transformer
  language models.
\newblock In \emph{Proceedings of the 2024 Conference of the North American
  Chapter of the Association for Computational Linguistics: Human Language
  Technologies (Volume 1: Long Papers)}, pp.\  7239--7252. Association for
  Computational Linguistics, 2024.

\bibitem[Potluru et~al.(2024)Potluru, Borrajo, Coletta, Dalmasso, El-Laham,
  Fons, Ghassemi, Gopalakrishnan, Gosai, Krea{\v{c}}i{\'c}, Mani, Obitayo,
  Paramanand, Raman, Solonin, Sood, Vyetrenko, Zhu, Veloso, and
  Balch]{potluru2024syntheticdataapplicationsfinance}
Potluru, V.~K., Borrajo, D., Coletta, A., Dalmasso, N., El-Laham, Y., Fons, E.,
  Ghassemi, M., Gopalakrishnan, S., Gosai, V., Krea{\v{c}}i{\'c}, E., Mani, G.,
  Obitayo, S., Paramanand, D., Raman, N., Solonin, M., Sood, S., Vyetrenko, S.,
  Zhu, H., Veloso, M., and Balch, T.
\newblock Synthetic data applications in finance.
\newblock \emph{arXiv preprint arXiv:2401.00081}, 2024.

\bibitem[Rajbhandari et~al.(2020)Rajbhandari, Rasley, Ruwase, and
  He]{zero_initialization}
Rajbhandari, S., Rasley, J., Ruwase, O., and He, Y.
\newblock {ZeRO}: Memory optimizations toward training trillion parameter
  models.
\newblock In \emph{SC20: International Conference for High Performance
  Computing, Networking, Storage and Analysis}, pp.\  1--16. IEEE, 2020.

\bibitem[Raman et~al.(2024)Raman, Ganesh, and Veloso]{raman2024scalable}
Raman, N., Ganesh, S., and Veloso, M.
\newblock Scalable representation learning for multimodal tabular transactions.
\newblock \emph{arXiv preprint arXiv:2410.07851}, 2024.

\bibitem[Ratliff-Crain et~al.(2025)Ratliff-Crain, Van~Oort, Koehler, and
  Tivnan]{ratliff2023revisiting}
Ratliff-Crain, E., Van~Oort, C.~M., Koehler, M. T.~K., and Tivnan, B.~F.
\newblock Revisiting {Cont}'s stylized facts for modern stock markets.
\newblock \emph{Quantitative Finance}, 25\penalty0 (9):\penalty0 1343--1373,
  2025.

\bibitem[Sirignano \& Cont(2021)Sirignano and Cont]{sirignano2021universal}
Sirignano, J. and Cont, R.
\newblock Universal features of price formation in financial markets:
  Perspectives from deep learning.
\newblock In \emph{Machine Learning and AI in Finance}, pp.\  5--15. Routledge,
  2021.

\bibitem[Touvron et~al.(2023)Touvron, Lavril, Izacard, Martinet, Lachaux,
  Lacroix, Rozi{\`e}re, Goyal, Hambro, Azhar, Rodriguez, Joulin, Grave, and
  Lample]{llama}
Touvron, H., Lavril, T., Izacard, G., Martinet, X., Lachaux, M.-A., Lacroix,
  T., Rozi{\`e}re, B., Goyal, N., Hambro, E., Azhar, F., Rodriguez, A., Joulin,
  A., Grave, E., and Lample, G.
\newblock {LLaMA}: Open and efficient foundation language models.
\newblock \emph{arXiv preprint arXiv:2302.13971}, 2023.

\bibitem[Vaswani et~al.(2017)Vaswani, Shazeer, Parmar, Uszkoreit, Jones, Gomez,
  Kaiser, and Polosukhin]{vaswani2017attention}
Vaswani, A., Shazeer, N., Parmar, N., Uszkoreit, J., Jones, L., Gomez, A.~N.,
  Kaiser, {\L}., and Polosukhin, I.
\newblock Attention is all you need.
\newblock In \emph{Advances in Neural Information Processing Systems},
  volume~30, 2017.

\bibitem[Vyetrenko et~al.(2020)Vyetrenko, Byrd, Petosa, Mahfouz, Dervovic,
  Veloso, and Balch]{vyetrenko2019realrealismmetricsrobust}
Vyetrenko, S., Byrd, D., Petosa, N., Mahfouz, M., Dervovic, D., Veloso, M., and
  Balch, T.~H.
\newblock Get real: Realism metrics for robust limit order book market
  simulations.
\newblock In \emph{Proceedings of the First ACM International Conference on AI
  in Finance}, pp.\  1--8, 2020.

\bibitem[Wen et~al.(2023)Wen, Zhou, Zhang, Chen, Ma, Yan, and
  Sun]{wen2022transformers}
Wen, Q., Zhou, T., Zhang, C., Chen, W., Ma, Z., Yan, J., and Sun, L.
\newblock Transformers in time series: A survey.
\newblock In \emph{Proceedings of the Thirty-Second International Joint
  Conference on Artificial Intelligence}, pp.\  6778--6786, 2023.

\bibitem[Wiese et~al.(2020)Wiese, Knobloch, Korn, and
  Kretschmer]{wiese2020quant}
Wiese, M., Knobloch, R., Korn, R., and Kretschmer, P.
\newblock Quant {GANs}: Deep generation of financial time series.
\newblock \emph{Quantitative Finance}, 20\penalty0 (9):\penalty0 1419--1440,
  2020.

\bibitem[Xiao et~al.(2025)Xiao, Ventre, Wang, Li, Huan, and Liu]{xiao2025lit}
Xiao, Y., Ventre, C., Wang, Y., Li, H., Huan, Y., and Liu, B.
\newblock {LiT}: Limit order book transformer.
\newblock \emph{Frontiers in Artificial Intelligence}, 8:\penalty0 1616485,
  2025.

\bibitem[Zhang et~al.(2019)Zhang, Zohren, and Roberts]{zhang2019deeplob}
Zhang, Z., Zohren, S., and Roberts, S.
\newblock {DeepLOB}: Deep convolutional neural networks for limit order books.
\newblock \emph{IEEE Transactions on Signal Processing}, 67\penalty0
  (11):\penalty0 3001--3012, 2019.

\end{thebibliography}

\newpage
\appendix
\section{Appendix}\label{sec:appendix/appendix_a}
\subsection{Foundation Models for Structured Data}

\begin{wrapfigure}{r}{0.5\textwidth}
    \centering
    \includegraphics[width=0.48\textwidth]{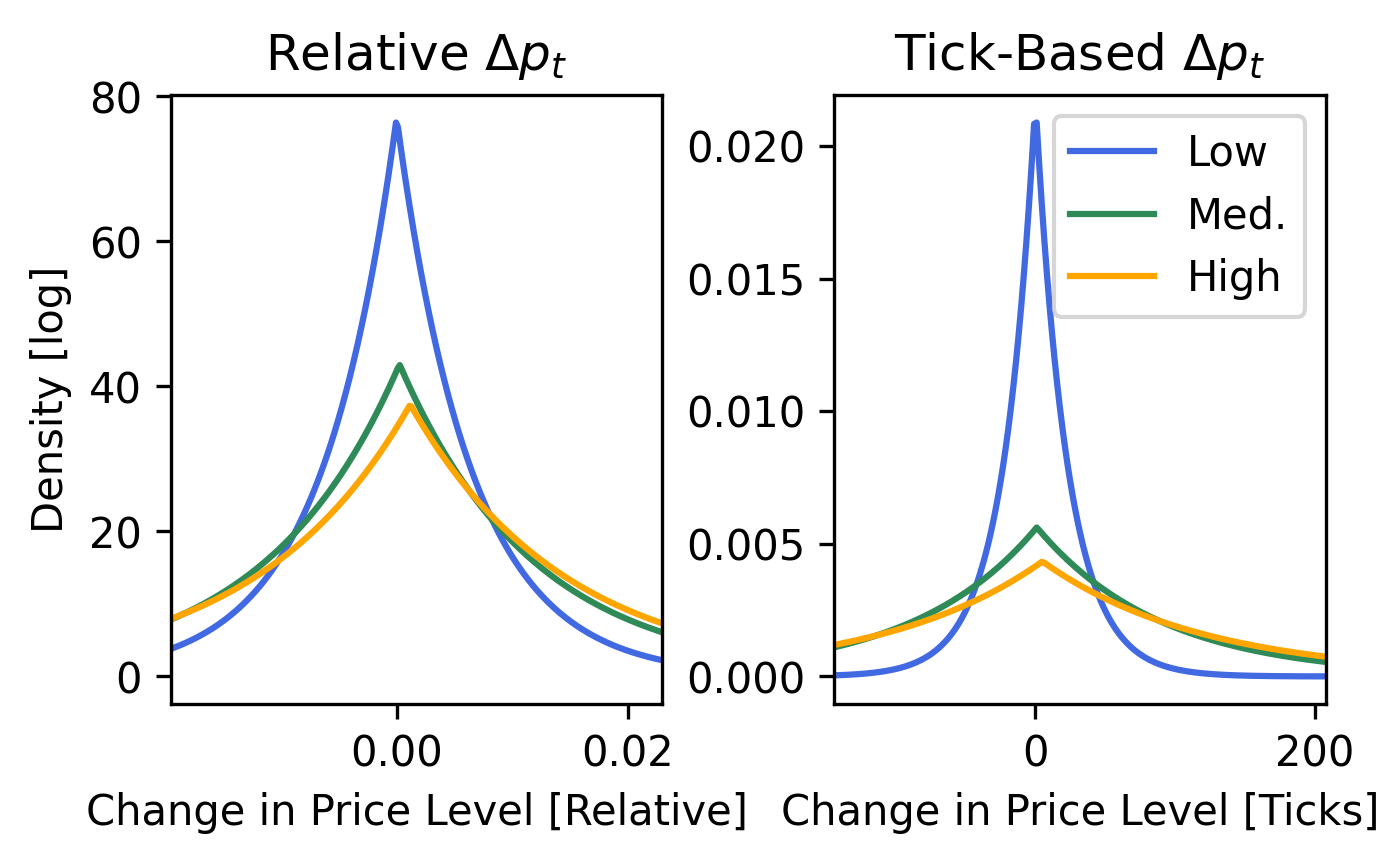}
    \caption{\textbf{Tick-based vs. relative features.} Properties of tick-based vs. relative feature construction for the sample feature $\Delta p_t$, across liquidity profiles. Relative features generalize better across assets than absolute, tick-based features.}
    \label{fig:tick_vs_relative}
\end{wrapfigure}

This work draws inspiration from the success of foundation models beyond NLP, in domains with structured sequential data. This paradigm has been successfully adapted by treating domain-specific sequences as a form of "language." In genomics, for example, models treat DNA or protein sequences as sentences to learn fundamental biological patterns~\citep{ji2021dnabert}. For general-purpose time-series forecasting, large pre-trained models have demonstrated strong zero-shot performance on unseen series~\citep{garza2023timegpt}. Similarly, for tabular data, Transformers pre-trained on a diverse collection of tables can perform inference on new, smaller tables without task-specific fine-tuning~\citep{badaro2023transformers, hollmann2025accurate}. In finance, by framing order flow as a structured language of market events, our approach aligns with this proven paradigm, arguing for its direct applicability to the unique challenges of financial data.

\subsection{The Transformer as a Natural Fit}
The Transformer architecture is uniquely suited to address these challenges. Its core components map naturally to the fundamental properties of order flow data:
\begin{itemize}[leftmargin=20pt]
    \item \textbf{Self-Attention}: The attention mechanism is designed to capture complex, long-range dependencies within a sequence. This makes it an ideal tool for modeling the long memory and intricate, non-linear interactions inherent in order flow.
    \item \textbf{Sequence-to-Sequence Framework}: As an autoregressive, sequence-based model, the Transformer inherently handles the asynchronous, event-driven nature of the data, where the time between events is itself a feature to be learned.
    \item \textbf{Adapting to Multi-feature Sequences}: While Transformers excel at processing univariate text, our trade events are multi-feature tuples. A key challenge is thus to effectively discretize and tokenize these mixed-type features into a processable sequence, which motivates our novel tokenization and embedding methodology.
\end{itemize}

We develop an end-to-end methodology to build TradeFM, a generative foundation model for market microstructure. The following four sections detail each component of our pipeline: our data processing and scale-invariant feature engineering (Section~\ref{sec:data_pipeline}), our universal tokenization scheme (Section~\ref{sec:tokenization}), the TradeFM model architecture (Section~\ref{sec:TradeFM}), and the closed-loop market simulator used for evaluation (Section~\ref{sec:market_simulator}).

\subsection{Mid-Price Estimation} \label{sec:appendix/midprice_estimation}

A robust estimate of the true market mid-price ($\midprice$) is critical for normalizing price-related features. While dedicated market data sources for this exist, they are often expensive. Given our access to raw transaction data, we seek to estimate this value directly. In our partial-information setting, we primarily observe the execution prices ($\execprice$) for consummated trades.
The raw stream of $\execprice$ is a noisy version of the true mid-price $\midprice$.

A naive approach, such as a simple rolling average of execution prices, is insufficient. A fixed-width window of trades is not comparable across assets with different liquidity levels; a 50-trade window may span less than a second for a highly liquid asset but several hours for an illiquid one. A time-based window (e.g., 2 seconds) is more relevant, but a simple average still fails to account for trade volume. For example, an average that gives equal weight to a 1,000-share trade at \$10.00 and a 1-share trade at \$9.00 would produce a misleading estimate.

The conventional solution to this problem is the volume-weighted average price (VWAP), which is 

\begin{equation}
    \hat{p}_t^{\text{VWAP}} = \frac{\sum_{i=0}^{W} v_{t-i} \execprice[t-i]}{\sum_{i=0}^{W} v_{t-i}}
\end{equation}

To make this estimator more reactive to recent information, we introduce \textbf{Exponentially-Weighted Volume-Weighted Average Price (EW-VWAP)}. This is calculated by maintaining two separate exponential moving averages (EMAs): one for the volume-weighted price and one for the volume itself. For each incoming trade with execution price $\execprice$ and volume $v_t$, we update the EMAs for the numerator ($N_t$) and denominator ($D_t$) as follows:
\begin{align*}
    N_t &= \alpha \cdot (\execprice \cdot v_t) + (1 - \alpha) \cdot N_{t-1} \\
    D_t &= \alpha \cdot v_t + (1 - \alpha) \cdot D_{t-1}
\end{align*}
The EW-VWAP at time $t$ is then the ratio of these two values:
\begin{equation}
\hat{p}_t^{\text{EW-VWAP}} = \frac{N_t}{D_t}
\end{equation}
The smoothing factor $\alpha$ is determined by a time-based halflife, ensuring that the estimate gives more weight to larger and more recent trades in a temporally consistent manner. This provides a stable and representative price benchmark that reflects the price at which the bulk of recent market activity has occurred.
This estimator is used throughout the paper to normalize all price-related features to a common scale.

\begin{figure*}[p]
    \centering
    \includegraphics[width=\linewidth]{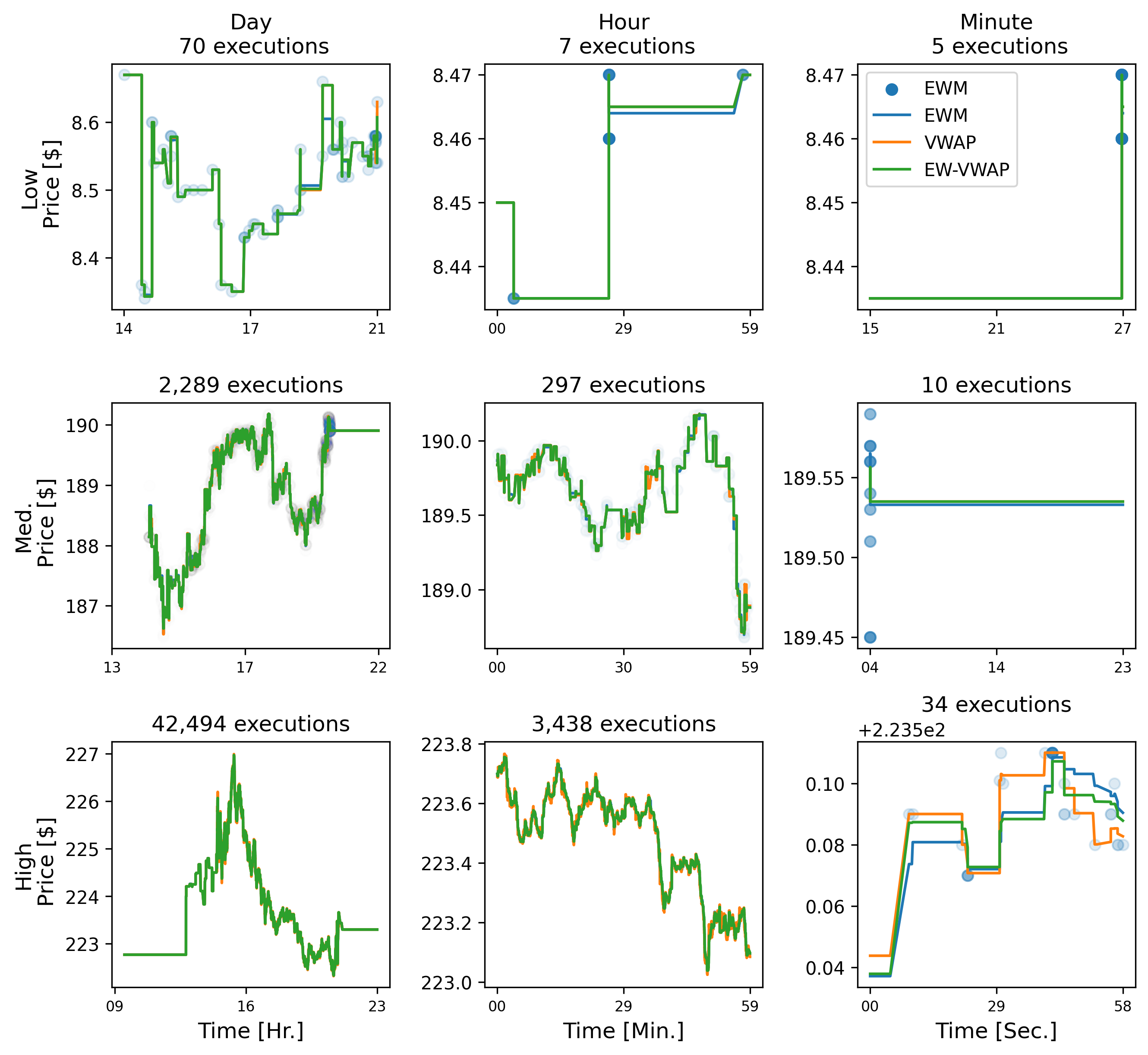}
    \caption{\textbf{Mid-price estimator comparison.} Our proposed EW-VWAP provides a more stable and responsive estimate than standard VWAP or EWM, closely tracking the executed fill prices across different time scales and liquidity regimes.}
    \label{fig:midprice_estimators}
\end{figure*}

\subsection{Tokenization Example} 
\label{sec:sample_encoding}

\begin{table*}
\centering
\begin{tabular}{c|c|c|r|r|c|c|r|r}
\toprule
\textbf{Time-} & \textbf{Time-} & \textbf{Asset} & \textbf{Avg. Daily} & \textbf{Midprice} & \textbf{Action} & \textbf{Side} & \textbf{Order} & \textbf{Vol.} \\
\textbf{step} & \textbf{stamp} & & \textbf{Vol. (shs)} & \textbf{(\$)} & & & \textbf{Price (\$)} & \textbf{(shs)} \\
\midrule
\multicolumn{9}{c}{$\vdots$} \\
42 & 09:45:30 & AAPL & 53,496,022 & 182.45 & ADD & BUY & 182.44 & 500 \\
43 & 09:45:38 & AAPL & 53,496,022 & 182.48 & ADD & SELL & 182.50 & 750 \\
44 & 09:45:52 & AAPL & 53,496,022 & 182.50 & CANCEL & BUY & 182.49 & 300 \\
\multicolumn{9}{c}{$\vdots$} \\
\bottomrule
\end{tabular}
\caption{\textbf{Example trade sequence.} Toy example of trading activity for an imaginary AAPL sequence, demonstrating the multi-feature and heterogeneous nature of our data pre-tokenization.}
\label{tab:trading_activity}
\end{table*}

Given the imaginary sequence of trade events $e_t$ constructed in Table~\ref{tab:trading_activity}, our features for timestep $t=43$ are as follows:

\begin{itemize}[leftmargin=20pt, nosep]
\item $\Delta t_t=w_t-w_{t-1}=\text{09:45:38}-\text{09:45:30}=8\text{sec}$
\item $\delta p_t=\frac{\orderprice-\midpriceest}{\midpriceest}=\frac{\$182.50-\$182.48}{\$182.48}=\frac{\$0.02}{\$182.48}=0.011\%=+1.1\,\text{bps}$
\item $v_t=750\text{shs}$
\item $a_t=\text{Add Order}$
\item $s_t=\text{Sell}$
\end{itemize}

Using our calibrated bins, we would discretize these features to the bin indices:
\begin{itemize}[leftmargin=20pt, nosep]
\item $i_{\Delta t_t}=\text{bin }11$
\item $i_{\delta p_t}=\text{bin }7$
\item $i_{v_t}=\text{bin }7$
\item $i_{a_t}=\text{bin }0$
\item $i_{s_t}=\text{bin }1$
\end{itemize}

Using Equation~\ref{eq:tokenization} would yield:

\begin{center}
    $i_{\text{trade}}=6011$
\end{center}

Assuming an opening price $p_0=\$179.50$, we would have change in price level feature:
\begin{center}
    $\Delta p_t=\frac{\midpriceest-p_0}{p_0}=\frac{\$182.48-\$179.50}{\$179.50}=1.66\%=+166\,\text{bps}$
\end{center}

Discretizing this using our bins would give $i_{\Delta p_t}=19$. Based on the asset's average daily volume of 53 million, which falls into the high liquidity bin, our liquidity indicator $i_l$ would be $2$. If this sequence was a market-level sequence, we would have market-participant indicator $I_{MP}=0$.

Our final model input would then be:
\begin{center}
    $[i_l, I_{MP}, i_{\Delta p_t}, i_{\text{trade}}]=[2,0,19,6011]$.
\end{center}

\begin{table}[htbp]
\centering
\begin{tabular}{l|ccccc}
\toprule
\textbf{Model Size} &
\makecell{\textbf{Num.}\\\textbf{Train Tokens}} &
\makecell{\textbf{Batch}\\\textbf{Size}} &
\textbf{GPUs} &
\makecell{\textbf{Train Time}\\\textbf{/ Epoch (hrs)}} &
\textbf{Optimization} \\
\midrule
125M  & 2.6B & 24 & 3xA100 & 17 & Accelerate \\
250M  & 6.4B & 32 & 4xA10G & 29 & DeepSpeed~\citep{zero_initialization} \\
500M  & 10.7B & 24 & 3xA100 & 128 & Accelerate \\
\bottomrule
\end{tabular}
\caption{\textbf{Training configuration.} Setup details for different model sizes, including token counts, batch sizes, hardware, and training time per epoch.}
\label{tab:model_training_config}
\end{table}

\section{Reproducibility Guide}\label{sec:appendix/reproducibility}
\subsection{Model Backbone}\label{sec:TradeFM/Model}
TradeFM is a decoder-only Transformer, trained from scratch with a custom configuration. The architecture is based on the Llama family~\citep{llama} and incorporates modern enhancements for efficiency and performance, including:
\begin{itemize}[leftmargin=20pt]
    \item \textbf{Grouped-Query Attention (GQA)}: Balances the performance of Multi-Head Attention with the reduced memory bandwidth of Multi-Query Attention, enabling faster inference and larger batch sizes.
    \item \textbf{Rotary Position Embeddings (RoPE)}: Encodes relative positional information by applying a rotation to query and key vectors, which has been shown to improve generalization for long sequences.
\end{itemize}

\subsection{Model Hyperparameters and Scaling}\label{sec:TradeFM/Hyperparams}
The model size is guided by the Chinchilla scaling laws, which prescribe a compute-optimal balance between model size and unique training tokens, implying approximately 20 tokens per parameter~\citep{kaplan_scaling, chinchilla_scaling}. Given our dataset of 10.7 billion unique tokens, this implies a target model size of around 525 million parameters. We note that our 4-epoch training schedule means the model sees approximately 42.8B total tokens, exceeding this ratio with repeated data; this follows recommendations for training in data-constrained regimes~\citep{scaling_data_constrained}. Our final hyperparameters are as follows:
\begin{itemize}[leftmargin=20pt, nosep]
    \item \textbf{Context Length}: 1,024 tokens
    \item \textbf{Hidden Layers}: 32
    \item \textbf{Embedding Dimension}: 1,024
    \item \textbf{Intermediate MLP Size}: 4,096
    \item \textbf{Attention Heads}: 32
    \item \textbf{Key-Value Heads (GQA)}: 8 heads, 4 groups
    \item \textbf{Total Parameters}: 524.4 Million
\end{itemize}

The 1:4 ratio between the embedding dimension and the intermediate MLP size is chosen in accordance with best practices for Transformer models~\citep{llm_depth_width}.

\subsection{Training Configuration}\label{sec:TradeFM/Training}
We train the model on an AWS instance with 3 Nvidia A100 GPUs, each with 80GB of RAM. All training is performed in \texttt{fp16} half-precision. To achieve an effective batch size of 4,032, we use a per-device batch size of 24 and gradient accumulation over 56 steps. For further memory optimization and training acceleration, we use the Accelerate library. The model is trained using the AdamW optimizer with a linear learning rate schedule, a learning rate of $5 \times 10^{-5}$, and 500 steps of warmup. Following recommendations for training on large datasets~\citep{scaling_data_constrained}, we train for a total of 4 epochs.

Due to compute constraints, different model sizes in memory, and different dataset sizes implied by Chinchilla scaling laws, our training setups vary slightly between model sizes. We summarize these variations in Table~\ref{tab:model_training_config}.

\begin{figure}[htbp]
    \centering
    \begin{subfigure}[t]{0.48\linewidth}
        \centering
        \includegraphics[width=\linewidth]{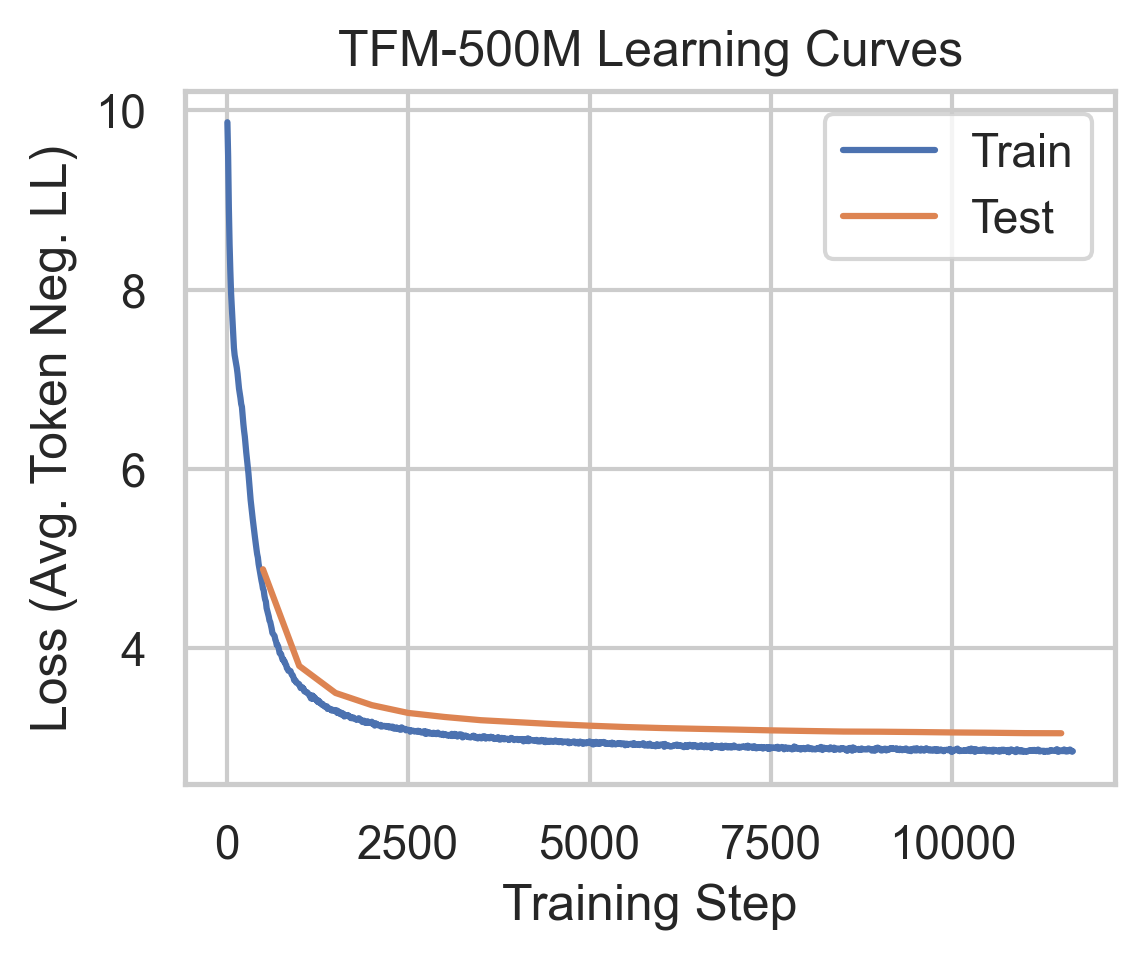}
        \caption{\textbf{TradeFM-500M learning curve.} Average token negative log likelihood over the test set constituting Jan. through Sept. 2025.}
        \label{fig:learning_curve}
    \end{subfigure}
    \hfill
    \begin{subfigure}[t]{0.48\linewidth}
        \centering
        \includegraphics[width=\linewidth]{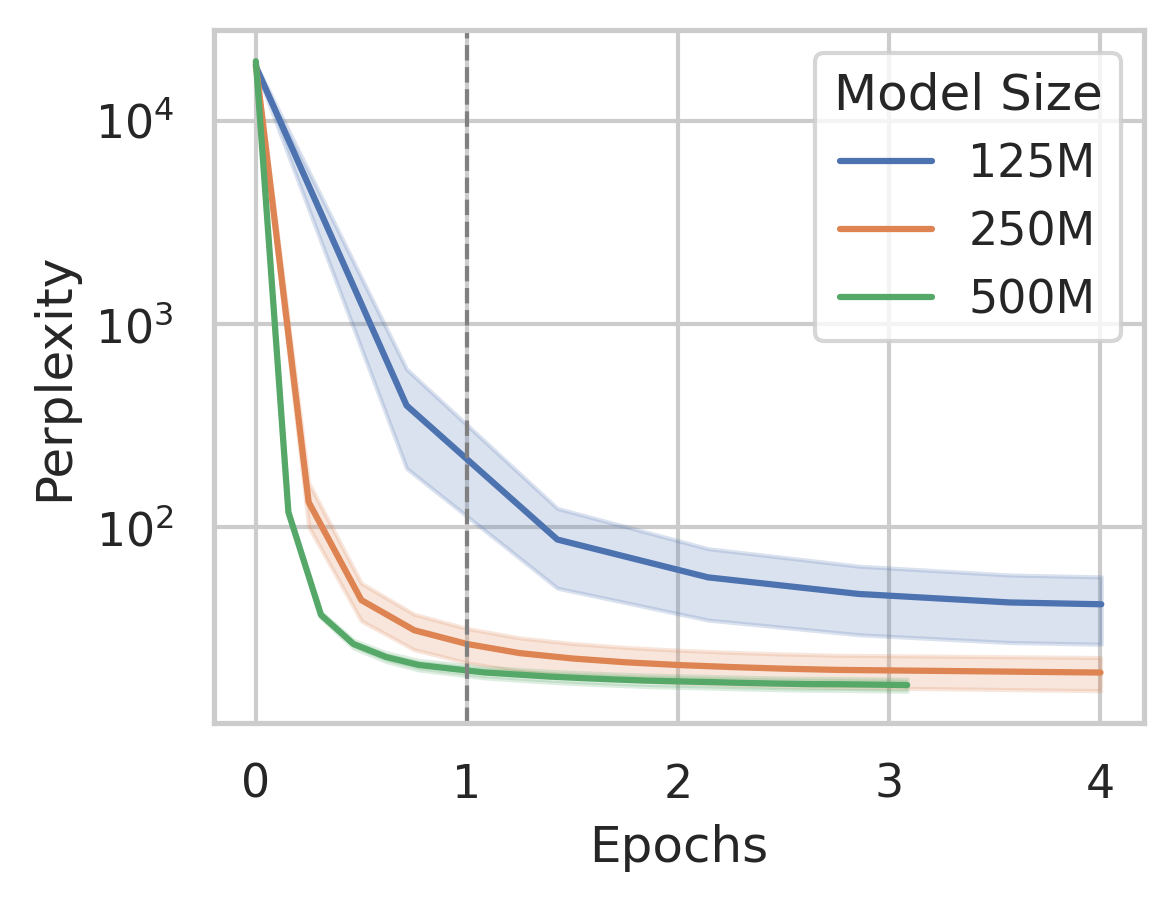}
        \caption{\textbf{Perplexity vs. epochs.} Out-of-sample perplexity for all TradeFM models. Epoch 1 represents one pass over the entire dataset; subsequent training is on repeated data.}
        \label{fig:ppl_vs_epochs}
    \end{subfigure}
    \caption{\textbf{Training dynamics.} (a) Learning curve for TradeFM-500M. (b) Out-of-sample perplexity across model sizes and epochs.}
    \label{fig:training_dynamics}
\end{figure}

\subsection{Tokenizer Pseudocode}
\begin{algorithm}[H]
\centering
\caption{Tokenizer Pseudocode}\label{alg:tokenizer}
\begin{algorithmic}[1]
\REQUIRE Flattened trading data with features: price, volume, time, action, side, participant
\REQUIRE Bin counts: $N_{\text{price}}$, $N_{\text{depth}}$, $N_{\text{volume}}$, $N_{\text{time}}$, $N_{\text{side}}=2$, $N_{\text{action}}=2$
\FOR{each feature $f$ in \{price, depth, volume, time\}}
    \STATE Remove NaN and infinite values from $f$
    \STATE Compute upper outlier threshold $u = \text{percentile}(f, 99)$
    \IF{feature is double-sided}
        \STATE Compute lower outlier threshold $l = \text{percentile}(f, 1)$
        \STATE Assign values outside $[l, u]$ to lower / upper outlier bins
    \ELSE
        \STATE Assign values above $u$ to upper outlier bin
    \ENDIF
    \IF{using equal-frequency bins}
        \STATE Compute bin edges $B_f$ using quantile binning with $N_f$ bins
    \ELSE
        \STATE Compute bin edges $B_f$ using histogram binning with $N_f$ bins
    \ENDIF
    \STATE Digitize $f$ into bin indices $I_f$ using $B_f$
    \STATE Handle outliers: assign out-of-range values to edge bins as needed
\ENDFOR

\FOR{each categorical feature $c$ in \{action, side, participant\}}
    \STATE Map each category to a unique integer index $I_c$
\ENDFOR

\FOR{each order $o$ in the dataset}
    \FOR{each feature $f$}
        \IF{$o[f]$ is NaN}
            \STATE Impute $o[f]$ with a random valid bin index or default value
        \ENDIF
    \ENDFOR
    \STATE Compute token index for $o$:
    \STATE $T_o = I_{\text{action}} \times N_{\text{side}} \times N_{\text{depth}} \times N_{\text{volume}} \times N_{\text{time}}$
    \STATE $\,\,\,\,\,\, + I_{\text{side}} \times N_{\text{depth}} \times N_{\text{volume}} \times N_{\text{time}}$
    \STATE $\,\,\,\,\,\, + I_{\text{depth}} \times N_{\text{volume}} \times N_{\text{time}}$
    \STATE $\,\,\,\,\,\, + I_{\text{volume}} \times N_{\text{time}}$
    \STATE $\,\,\,\,\,\, + I_{\text{time}}$
    \STATE Assign $T_o$ to order $o$
\ENDFOR
\end{algorithmic}
\end{algorithm}

\subsection{Market Simulation Pseudocode}
\begin{algorithm}[H]
\centering
\caption{Market Simulator: Part 1 - Initialization and Utilities}\label{alg:sim_part_1}
\begin{algorithmic}[1]
\REQUIRE Sequence of order transactions, initial price $p_0$, simulation parameters

\STATE \textbf{Initialize Exchange:}
    \STATE Set initial price $p_0$
    \STATE Initialize order book, midprice, fills, deletes, spreads, bid/ask volumes

\vspace{0.5em}
\STATE \textbf{Function:} \textsc{InitializeOrderBook}(order\_columns)
    \STATE Reset order book, midprice, fills, deletes, spreads, bid/ask volumes
    \STATE Set initial bid/ask to $p_0$

\vspace{0.5em}
\STATE \textbf{Function:} \textsc{GetOrderPrice}(transaction)
    \IF{order is market}
        \IF{side is Sell}
            \STATE price $\gets$ lowest ask
        \ELSE
            \STATE price $\gets$ highest bid
        \ENDIF
    \ELSE
        \STATE price $\gets$ (order price depth / 10,000) $\times$ current midprice $+$ current midprice
    \ENDIF
    \STATE \textbf{Return} price 

\vspace{0.5em}
\STATE \textbf{Function:} \textsc{GenerateFill}(best\_past\_order, order, quantity)
    \STATE Compute time since best\_past\_order
    \STATE Determine match price:
        \IF{both orders are market}
            \STATE price $\gets$ current midprice
        \ELSIF{order is market}
            \STATE price $\gets$ best\_past\_order price
        \ELSIF{best\_past\_order is market}
            \STATE price $\gets$ order price
        \ELSE
            \STATE price $\gets$ best\_past\_order price
        \ENDIF
    \STATE \textbf{Return} fill record with IDs, sides, prices, depths, volume, time delta
\end{algorithmic}
\end{algorithm}

\vspace{0.5em}
\begin{algorithm}[H]
\centering
\caption{Market Simulator: Part 2 - Simulation Step Functions}\label{alg:sim_part_2}
\begin{algorithmic}
\STATE \textbf{Function:} \textsc{StepOrderBook}(order)
    \STATE Extract side and price from order
    \WHILE{order volume $>$ 0}
        \STATE Find matching opposite-side orders based on price-time priority
        \IF{no matching orders}
            \STATE \textbf{break}
        \ENDIF
        \STATE Select best matching order (highest bid or lowest ask)
        \IF{best\_past\_order volume $>$ order volume}
            \STATE Reduce best\_past\_order volume by order volume
            \STATE Record fill and \textbf{return}
        \ELSIF{best\_past\_order volume $<$ order volume}
            \STATE Reduce order volume by best\_past\_order volume
            \STATE Remove best\_past\_order from book
            \STATE Record fill
        \ELSE
            \STATE Record fill
            \STATE Remove best\_past\_order from book
            \STATE \textbf{return}
        \ENDIF
    \ENDWHILE
    \IF{order volume $>$ 0}
        \STATE Add partially filled order to book
    \ENDIF

\vspace{0.5em}
\STATE \textbf{Function:} \textsc{StepMidprice}(transaction)
    \IF{transaction is Delete}
        \STATE Use current order book
    \ELSE
        \STATE Add transaction to temporary order book
    \ENDIF
    \STATE Update highest bid and lowest ask from book
    \STATE Compute midprice as average of highest bid and lowest ask
    \STATE Record midprice and bid/ask volumes

\vspace{0.5em}
\STATE \textbf{Function:} \textsc{StepSim}(transaction)
    \STATE Update transaction midprice
    \IF{action is Add}
        \STATE Compute order price
        \STATE Update midprice
        \STATE Step order book
    \ELSIF{action is Delete}
        \STATE Match on order ID
        \STATE Remove matching orders and record deletes
        \STATE Update midprice
    \ENDIF
    \STATE Record simulation time for profiling

\vspace{0.5em}
\STATE \textbf{Function:} \textsc{RunSimulation}(data)
    \STATE Initialize order book
    \FOR{each transaction in data}
        \STATE StepSim(transaction)
    \ENDFOR
    \STATE \textbf{Return} fills and midprice history

\end{algorithmic}
\end{algorithm}

\subsection{Zero-Intelligence Baseline} \label{sec:appendix/zi_baseline}

The Zero-Intelligence (ZI) agent is a canonical null model used to test whether a model learns complex, conditional dynamics beyond the market's basic structural properties~\citep{zi_traders, farmer2005predictive}. To provide a fair baseline, our ZI agent generates orders stochastically by sampling from distributions calibrated to match the marginals of key features in a 450-million-trade sample of the training data.

Specifically, side and action type are sampled from their empirical categorical distributions; interarrival time is sampled from a fitted Exponential distribution; order volume from a fitted Exponential distribution; and price depth is drawn from a Gaussian Mixture Model (GMM).

The resulting ZI agent orders are processed through the identical market simulator and evaluation pipeline as TradeFM to ensure a direct and fair comparison. We compute 2,048 rollouts of 1,024 events, and compute the same stylized facts.

\subsection{Compound Hawkes Baseline} \label{sec:appendix/hawkes_baseline}
Hawkes Processes are commonly applied to market data for their ability to robustly model interarrival times of self-exciting events \citep{bacry2015hawkes, jain_hawkes}. We adopt the Compound Hawkes model which combines a Hawkes process for modeling interarrival times with empirical distributions for modeling additional event features like volume and price depth. We use the same 450-million-trade data as is used to train our zero-intelligence baseline, and separate the data based on action and side.

We then fit a Hawkes process using a sum of exponential kernel, with 4 dimensions, one for each combination of action and side (buy-delete, buy-add, sell-delete, sell-add). For each of these action-side combinations we calibrate a Gaussian Mixture Model for price depths, and an Exponential for volume.

\section{Scaling Analysis} \label{sec:appendix/scaling}

\begin{figure}[htbp]
    \centering
    \includegraphics[width=\linewidth]{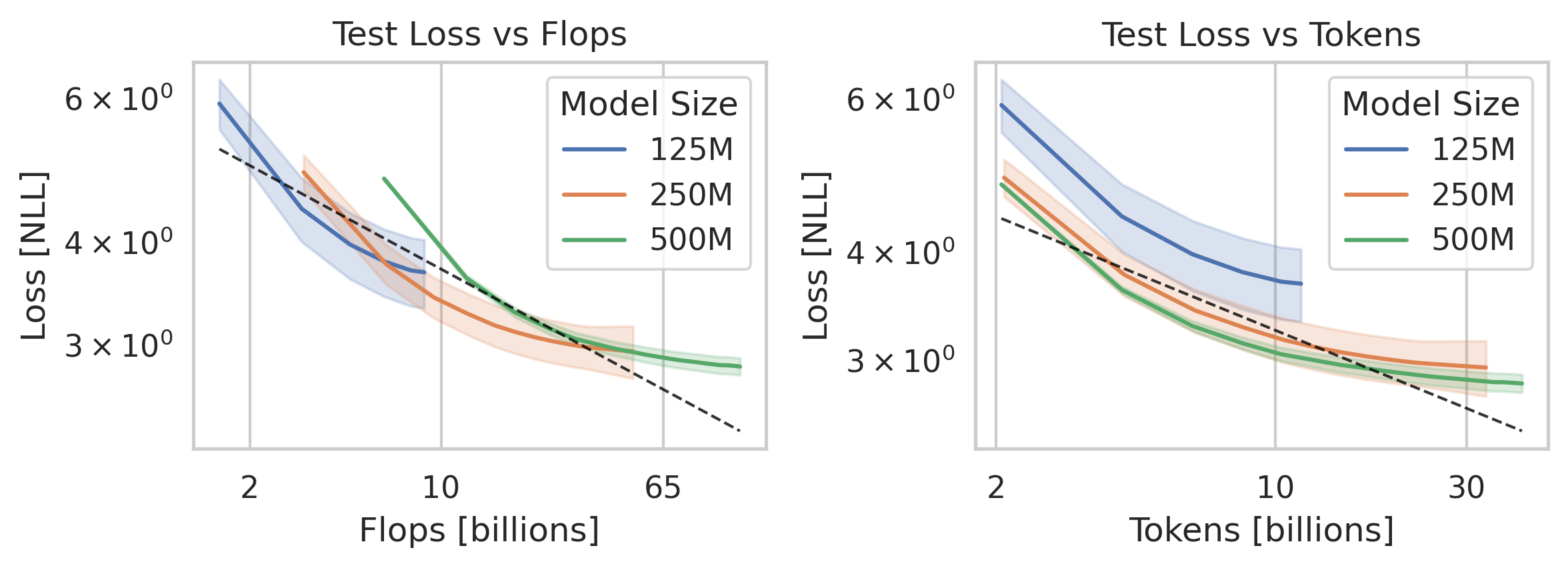}
    \vspace{-15pt}
    \caption{\textbf{Scaling law analysis.} Test loss (negative log likelihood) on held-out data one month in advance of the training data cutoff. The black dashed line represents the power law fit to the minimum loss frontier.}
    \label{fig:scaling_law_epochs}
    \vspace{5pt}
\end{figure}

To substantiate the \textit{Foundation Model} claim, we conducted a comprehensive scaling analysis of our approach. We trained models at three scales: approximately 125M, 250M, and 500M parameters.

Our \textbf{500M parameter model} is our largest model and serves as the primary evaluation target throughout this paper.
The scaling law plots in Figure~\ref{fig:scaling_law_epochs} demonstrate expected power-law relationships between compute, dataset size, and test loss. These plots include repeated data, as we train for four epochs. We verify this by computing the minimum loss frontier for compute and dataset size, fitting power laws to find that the test loss $L(C)$ with respect to compute in Flops $C$, and $L(D)$ with respect to dataset size in tokens $D$, follow:
\[L(C) \propto C^{-\alpha_C},\quad \alpha_C\approx0.18\]
\[L(D) \propto D^{-\alpha_D},\quad \alpha_D\approx0.19.\]

While 500M is small relative to general-purpose LLMs, it is large for the Financial Microstructure domain, and comparable in scale to other domain-specific models such as MaRS (${\sim}$1B parameters). Standard SOTA models in this field typically have $<$10M parameters (e.g., DeepLOB (60K params) and LOBS5 (6.3M params)). TradeFM represents a \textbf{$>$50$\times$ increase in model capacity} over existing domain-specific baselines.

\section{Extended Experimental Results}\label{sec:appendix/extended_results}

\subsection{Geographic Zero-Shot Generalization}
\label{sec:appendix/distribution_drift}

We evaluate the model, trained exclusively on US equities, on a hold-out set of assets from APAC markets (China and Japan). We detail these held-out datasets in Table~\ref{tab:geography_dataset_table}. Figure~\ref{fig:geographic_ood} shows the distribution of perplexity scores for one month of data (held out for all geographies), with significant overlap between US and APAC. The minimal degradation in perplexity on unseen markets confirms TradeFM's generalization capabilities.

\begin{table}[htbp]
\centering
\begin{tabular}{lccc}
\toprule
\textbf{Country} & \textbf{Number of Assets} & \textbf{Date-Asset Pairs} & \textbf{Tokens} \\
\midrule
US    & 6,885  & 81,203  & 857,017,219 \\
China & 4,926  & 68,925  & 37,408,529  \\
Japan & 2,932  & 37,235  & 286,476,052 \\
\bottomrule
\end{tabular}
\caption{\textbf{Geographic held-out datasets.} Dataset statistics for US, China, and Japan. All geographies are evaluated on Jan. 2025 data.}
\label{tab:geography_dataset_table}
\end{table}

\subsection{Simulator Validation} \label{sec:sim_validation}
\begin{figure}[htbp]
    \centering
    \includegraphics[width=\linewidth]{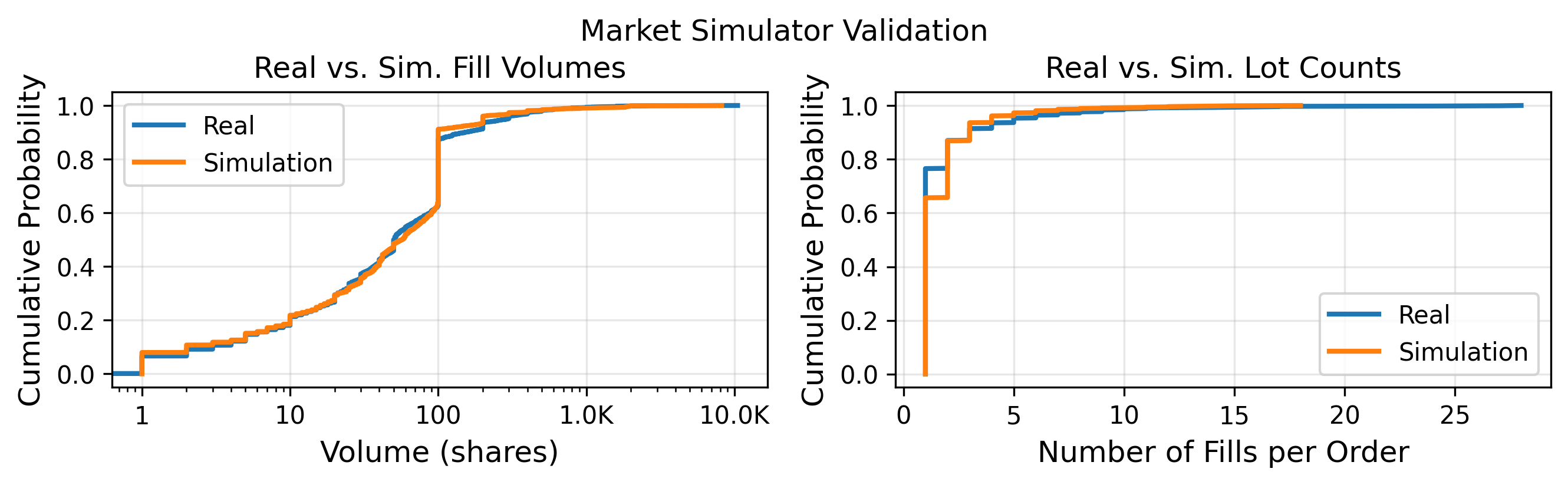}
    \caption{\textbf{Simulator fill validation.} Stylized facts of market simulator fills: (left) fill volumes; (right) lot counts. Correlations between simulated and actual fills are 0.91 for volumes and 0.98 for lot counts.}
    \label{fig:sim_stylized_facts}
\end{figure}

In order to evaluate the simulator, we replay sequences of real orders through it and compare the statistical properties of the resulting simulated trade fills against the real fills from our historical data. We focus on two key metrics: the cumulative distribution function (CDF) of fill volumes and the CDF of lot counts (the number of discrete fills required to complete a single order). As shown in Figure~\ref{fig:sim_stylized_facts}, we find a strong correspondence between the real and simulated distributions across assets of varying liquidity, confirming that our simulator is a high-fidelity environment for evaluation. We find correlations of 0.91 and 0.98 between sim and real volumes and lot counts, respectively.

\subsection{Temporal Drift}\label{sec:temporal_drift}

As financial markets are dynamic and market regimes are constantly changing, we investigate the tendency of model performance to drift over time. Our tokenizer's main contribution is to standardize representations of market features over both the liquidity and time regime. 

\begin{figure}[htbp]
    \centering
    \includegraphics[width=\linewidth]{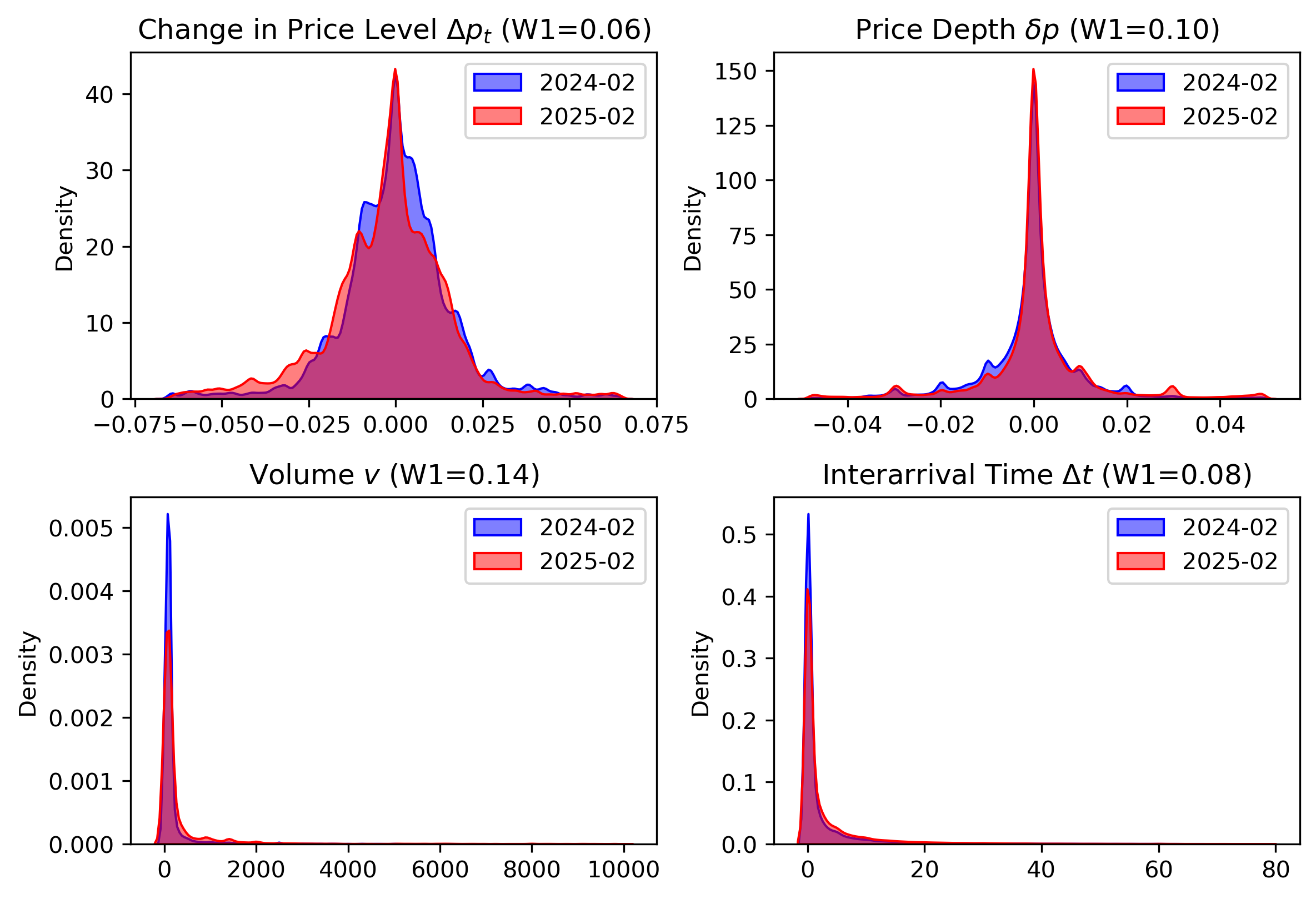}
    \caption{\textbf{Feature stationarity over time.} Kernel-density estimation of feature distributions from the tokenizer calibration period of Feb. 2024 to one year later in Feb. 2025. Our feature engineering successfully makes these features stationary, enabling generalization to out-of-distribution temporal regimes.}
    \label{fig:tokenizer_drift}
\end{figure}

\begin{figure}[htbp]
    \centering
    \includegraphics[width=\linewidth]{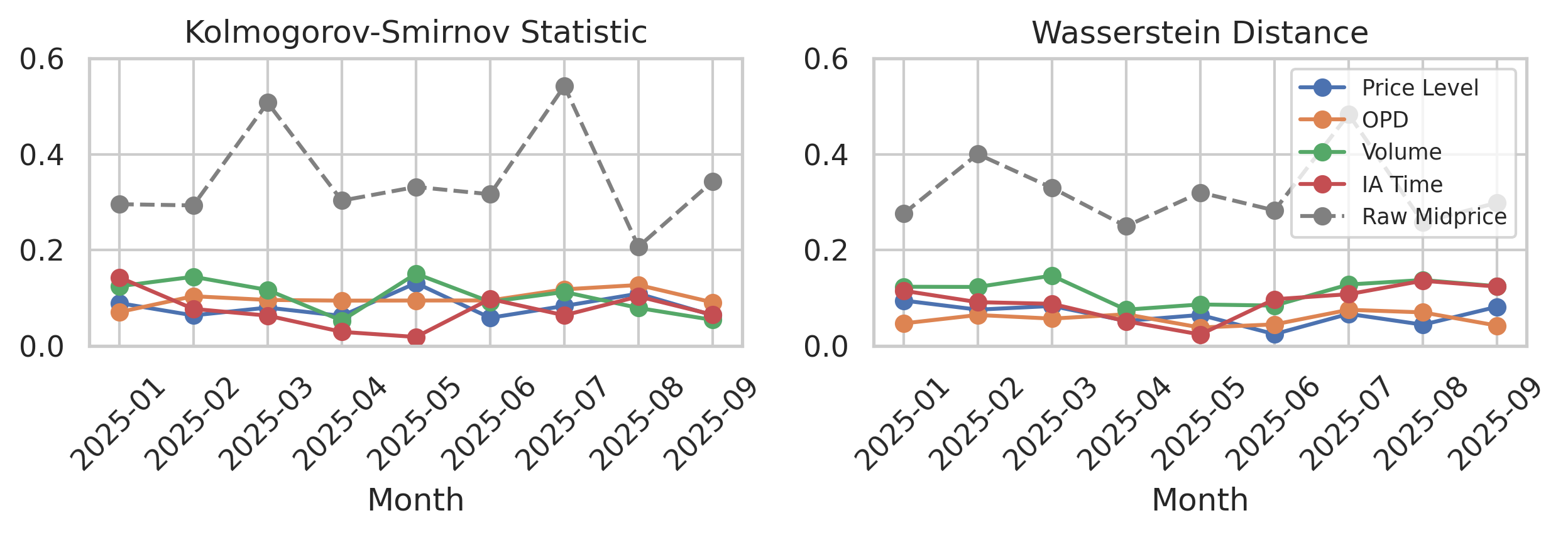}
    \caption{\textbf{Tokenizer drift over months.} Kolmogorov-Smirnov and Wasserstein distances between feature distributions during the tokenizer calibration month and held-out months. Raw mid-price, a non-stationary feature, is included for context. (OPD and IA Time in the legend correspond to Price Depth and Interarrival Time, respectively.)}
    \label{fig:tokenizer_drift_by_month}
\end{figure}

In Figure~\ref{fig:tokenizer_drift} we demonstrate the universality of these features by exploring the distribution of our relative price level, relative price depth, interarrival time, and volume features in both the month used to calibrate our tokenizer, Feb. 2024, and one year later in Feb. 2025. We observe that our features are stationary over this period even as volatility, price level, and other market conditions vary. Figure~\ref{fig:tokenizer_drift_by_month} shows the Kolmogorov-Smirnov and Wasserstein distance of each of these features between the tokenizer calibration month and each of 9 held-out months. We include a non-stationary feature, raw mid-price, to contextualize the stationarity of these metrics.

\begin{figure}[htbp]
    \centering
    \includegraphics[width=\linewidth]{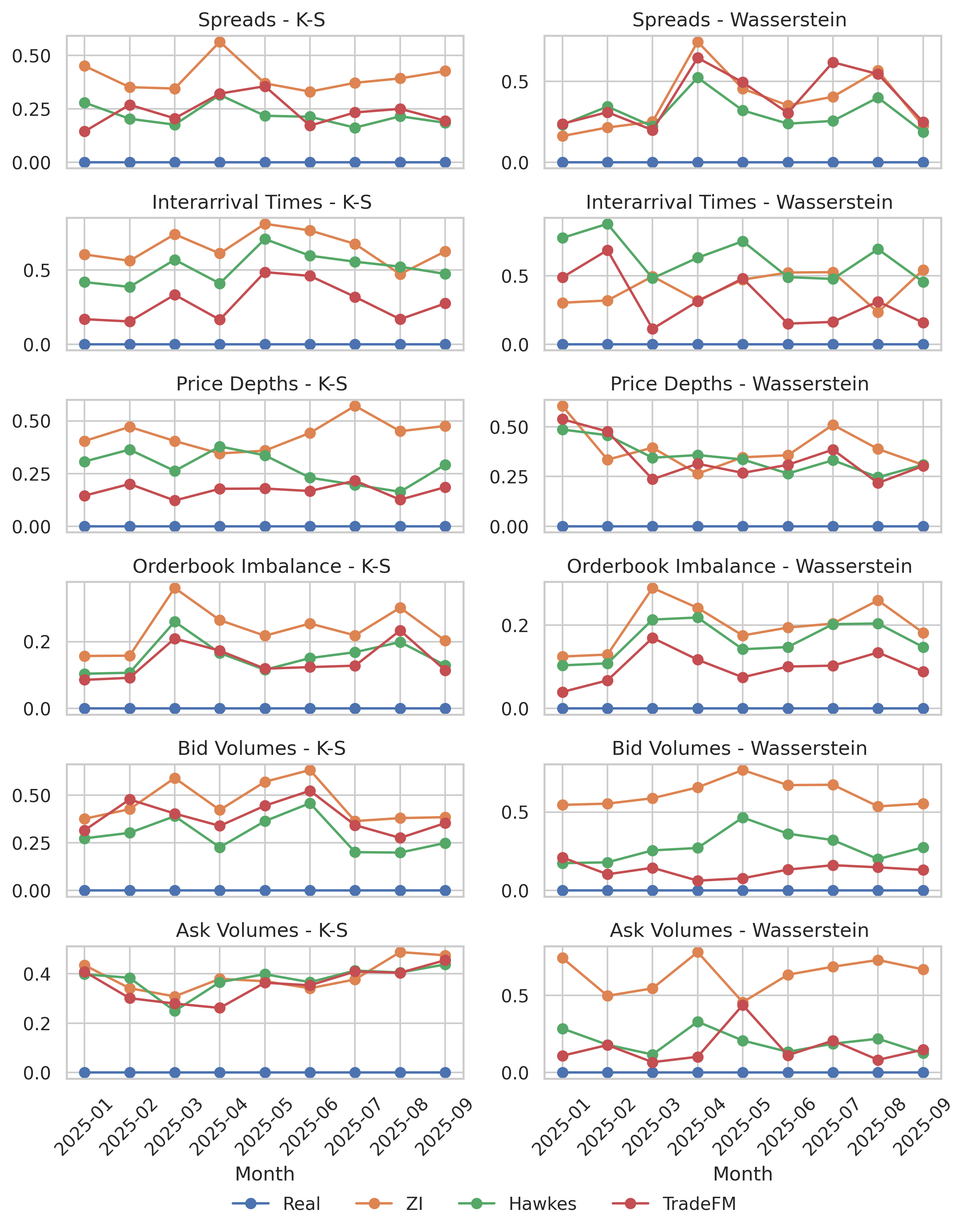}
    \caption{\textbf{Distributional fidelity over time.} Wasserstein distance and Kolmogorov-Smirnov statistic of feature distributions and emergent market factors from various methods over nine held-out months.}
    \label{fig:obi_over_time}
\end{figure}

In Figure~\ref{fig:obi_over_time} we extend the aggregated results in Table~\ref{tab:order_flow_distances} for all quantities of interest. We observe that while these metrics do vary within a range, the variance is small and our method mostly achieves higher fidelity than baselines.

\subsection{Synthetic Data Generation} \label{sec:synthetic_data}
\begin{figure}[htbp]
    \centering
    \includegraphics[width=0.65\textwidth]{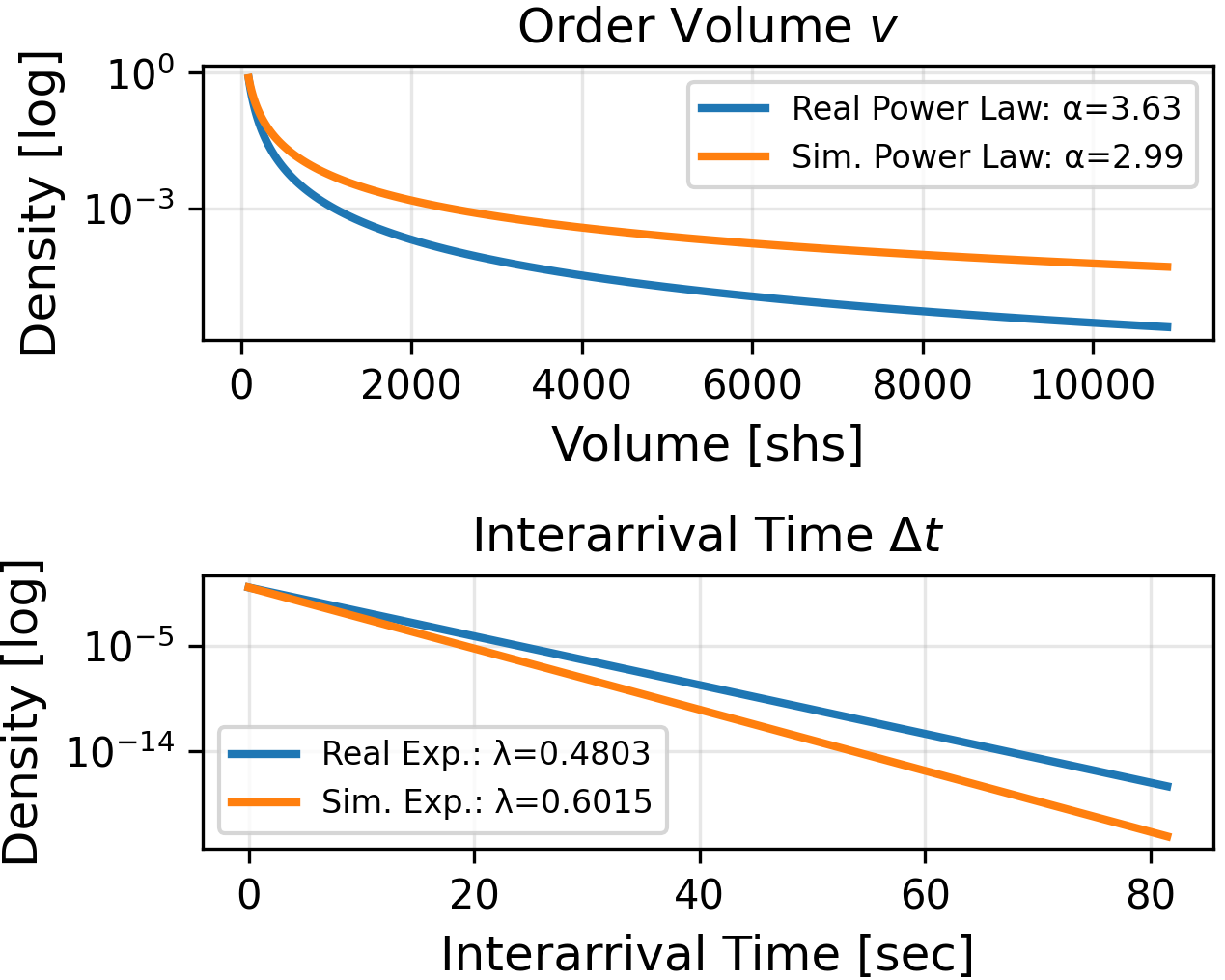}
    \caption{\textbf{Order-level distributional fidelity.} (Top) Simulated order volumes follow the power-law observed in real data. (Bottom) Simulated interarrival times match the exponential distribution of real data.}
    \label{fig:order_stylized_facts}
\end{figure}

TradeFM can serve as an engine for generating high-fidelity, privacy-preserving synthetic financial data.
The realism of the generated data is validated not only at the level of emergent price dynamics (Figure~\ref{fig:log_return_stylized_facts}), but also at the granular level of individual orders.
As shown in Figure~\ref{fig:order_stylized_facts}, the distributions of simulated order volumes and interarrival times closely match the power-law and exponential distributions observed in real data, respectively.
This high-fidelity generation is valuable for:
\begin{itemize}[leftmargin=20pt]
    \item \textbf{Backtesting Trading Strategies}: Synthetic data allows for robust testing of trading algorithms against a wide range of plausible market scenarios, reducing the risk of overfitting to a single historical path~\citep{potluru2024syntheticdataapplicationsfinance}.
    \item \textbf{Augmenting Sparse Datasets}: For illiquid assets where historical data is sparse, the model can generate additional data to facilitate more robust analysis and model training.
    \item \textbf{Privacy-Preserving Data Sharing}: The model can generate realistic datasets for academic or public use without revealing sensitive, proprietary transaction information.
\end{itemize}

\subsection{Market Simulation \& Stress Testing} \label{sec:market_sim_stress_test}
\begin{figure*}[htbp]
    \centering
    \includegraphics[width=\linewidth]{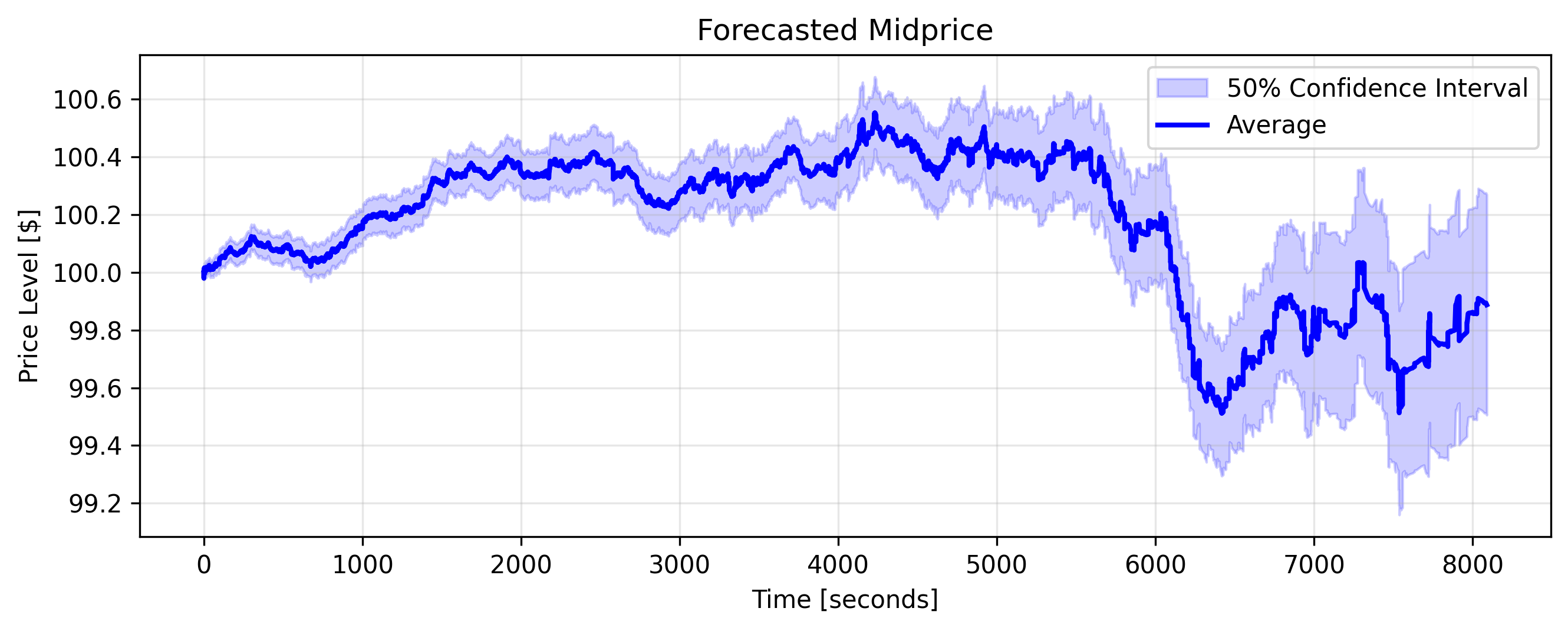}
    \vspace{-10pt}
    \caption{\textbf{Multi-step mid-price forecast.} Rollout-based forecast for an imaginary asset initially priced at \$100. The average trajectory and 50\% confidence interval over 256 simulations show the model generates plausible, non-stationary market paths.}
    \vspace{-10pt}
    \label{fig:midprice_traj}
\end{figure*}

The integrated TradeFM-simulator system functions as a high-fidelity environment for complex ``what-if'' analyses and stress testing. This allows for the study of systemic risk and market stability in a controlled environment. The ability to generate plausible, multi-step forecasts of future market trajectories, as illustrated in Figure~\ref{fig:midprice_traj}, is a direct outcome of this closed-loop simulation capability.

\begin{figure*}[htbp]
    \centering
    \includegraphics[width=\linewidth]{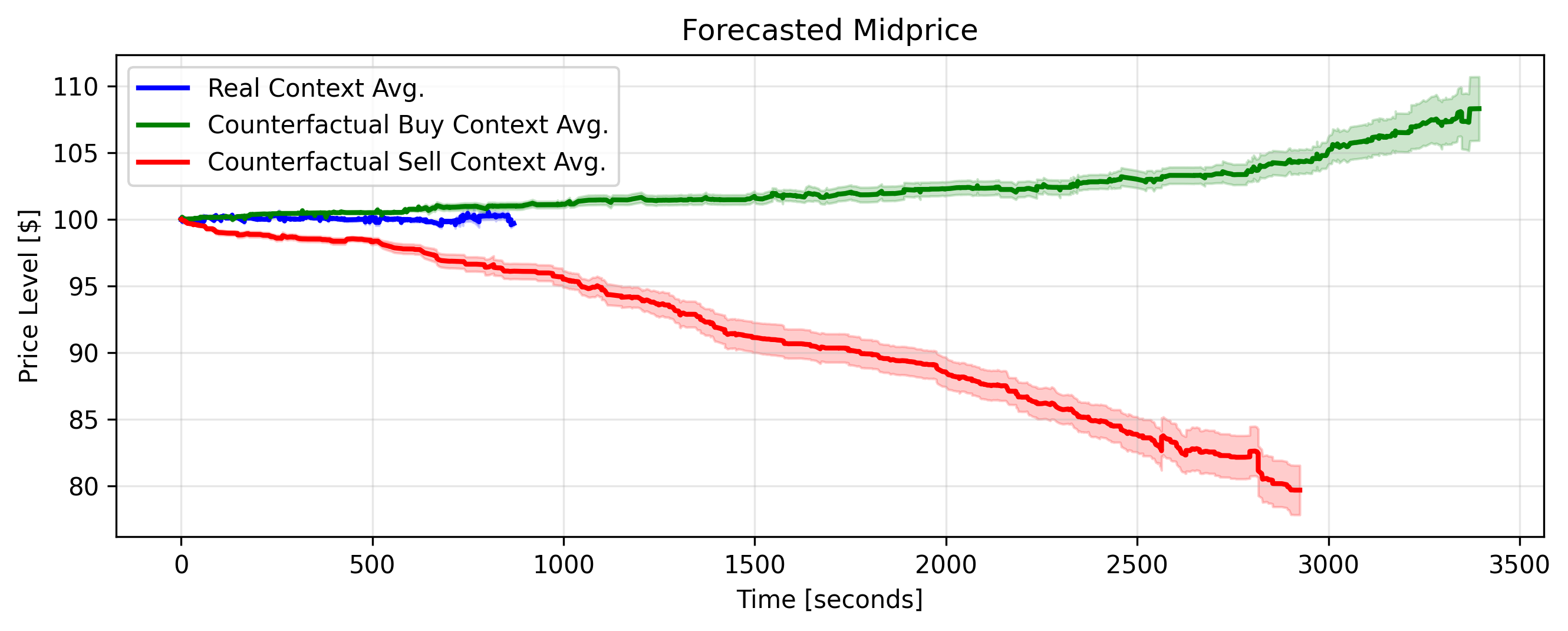}
    \vspace{-10pt}
    \caption{\textbf{Counterfactual stress testing.} The model's generated price path responds realistically to injected anomalous order flow (10x normal frequency), demonstrating its utility for impact analysis.}
    \label{fig:counterfactual_traj}
\end{figure*}

Such systems are also useful to regulators and risk managers~\citep{abides_economist}, who can use this system to simulate the market's response to extreme or counterfactual scenarios, such as by injecting large, anomalous orders into a historical context and observing the resulting price trajectory. Figure~\ref{fig:counterfactual_traj} demonstrates this capability -- for a sample asset, we artificially inject buy or sell orders at 10x the frequency found in the real context, and average mid-price forecasts over 10 rollouts. When we inject artificial sell orders, the mid-price drops, and when we inject buy orders, the mid-price rises, illustrating realistic behavior.

\subsection{Multi-Agent Modeling \& RL Fine-Tuning} \label{sec:multi_agent_modeling}
Our system provides a high-fidelity, interactive environment for training and evaluating sophisticated, learning-based agents. The pre-trained TradeFM can serve as a realistic ``background'' market, generating plausible and reactive counterparty order flow. This creates a dynamic training ground for:
\begin{itemize}[leftmargin=20pt]
    \item \textbf{Reinforcement Learning (RL) for Optimal Execution}: RL agents can be trained to learn optimal strategies for executing large orders by interacting with the simulated market, minimizing costs such as price impact and the bid-ask spread.
    \item \textbf{Multi-Agent Systems (MAS)}: The simulator can be populated with multiple, heterogeneous learning-based agents to study the emergent, collective behaviors and potential instabilities that arise from their interactions. The participant-level conditioning of our model provides a natural and powerful mechanism for initializing and fine-tuning diverse agent policies within such a system.
\end{itemize}

\end{document}